\documentclass[acmtog, nonacm, authorversion]{acmart}
\usepackage{pifont}
\usepackage{subcaption}
\usepackage{stfloats}
\usepackage{multirow} 
\usepackage{xcolor}
\usepackage[ruled]{algorithm2e} 

\usepackage{enumitem}

\SetAlFnt{\small}
\SetAlCapFnt{\small}
\SetAlCapNameFnt{\small}
\SetAlCapHSkip{0pt}

\begin{document}

\title{BoxCtrl: 3D-Aware Visual Prompting for Geometric Image Editing}

\author{Feifei Wang}
\affiliation{%
 \institution{City University of Hong Kong}
 \city{Hong Kong}
 \country{China}}
\email{feifwang3-c@my.cityu.edu.hk}

\author{Shiyuan Yang}
\affiliation{%
     \institution{City University of Hong Kong}
     \city{Hong Kong}
     \country{China}
}
\email{s.y.yang@my.cityu.edu.hk}

\author{Xiaoyu Li}
\affiliation{%
    \institution{Tencent}
    \city{Shenzhen}
    \country{China}}
\email{xliea@connect.ust.hk}
 
\author{Jing Liao}
\authornote{Corresponding author.}
\affiliation{%
     \institution{City University of Hong Kong}
     \city{Hong Kong}
     \country{China}}
\email{jingliao@cityu.edu.hk}

\begin{abstract}
As instruction-based editing models and multimodal large language models advance, diverse image editing tasks have become feasible. However, achieving precise and consistent geometric image editing, such as translating, scaling, and rotating in 3D space, remains a major challenge. In this work, we introduce \textbf{BoxCtrl}, a 3D-aware visual prompting framework. Unlike text-only or coarse 2D-guided approaches, our method introduces informative RGB 3D bounding boxes projected onto 2D images as visual prompts. The three orthogonal faces of each box are painted with distinct RGB colors, simultaneously encoding position, size, and orientation to provide a compact, intuitive in-context visual example. The key to BoxCtrl's success lies in these well-designed bounding boxes, which decouple geometric control from appearance control. This enables the model to learn consistent correspondences between faces of the same color in the latent space, leading to a precise understanding of geometric intentions and accurate editing results. We introduce a two-stage training paradigm: Supervised Fine-Tuning (SFT) followed by Reinforcement Learning (RL). To address paired data scarcity, we construct a large-scale synthetic dataset for SFT, equipping the model with fundamental editing capabilities. To bridge the synthetic-to-real domain gap, we incorporate an online RL stage leveraging unpaired real-world data. Guided by a reward function evaluating geometric accuracy and visual fidelity, our SFT-RL strategy significantly enhances geometric precision while maintaining photorealistic quality. Extensive experiments demonstrate that BoxCtrl achieves state-of-the-art performance across translation, rotation, scaling, and composite editing tasks. Our code is available at \url{https://github.com/beaglew/BoxCtrl}.

\end{abstract}

%
%

%
%


\begin{teaserfigure}
    \centering
    \includegraphics[width=1.0\textwidth]{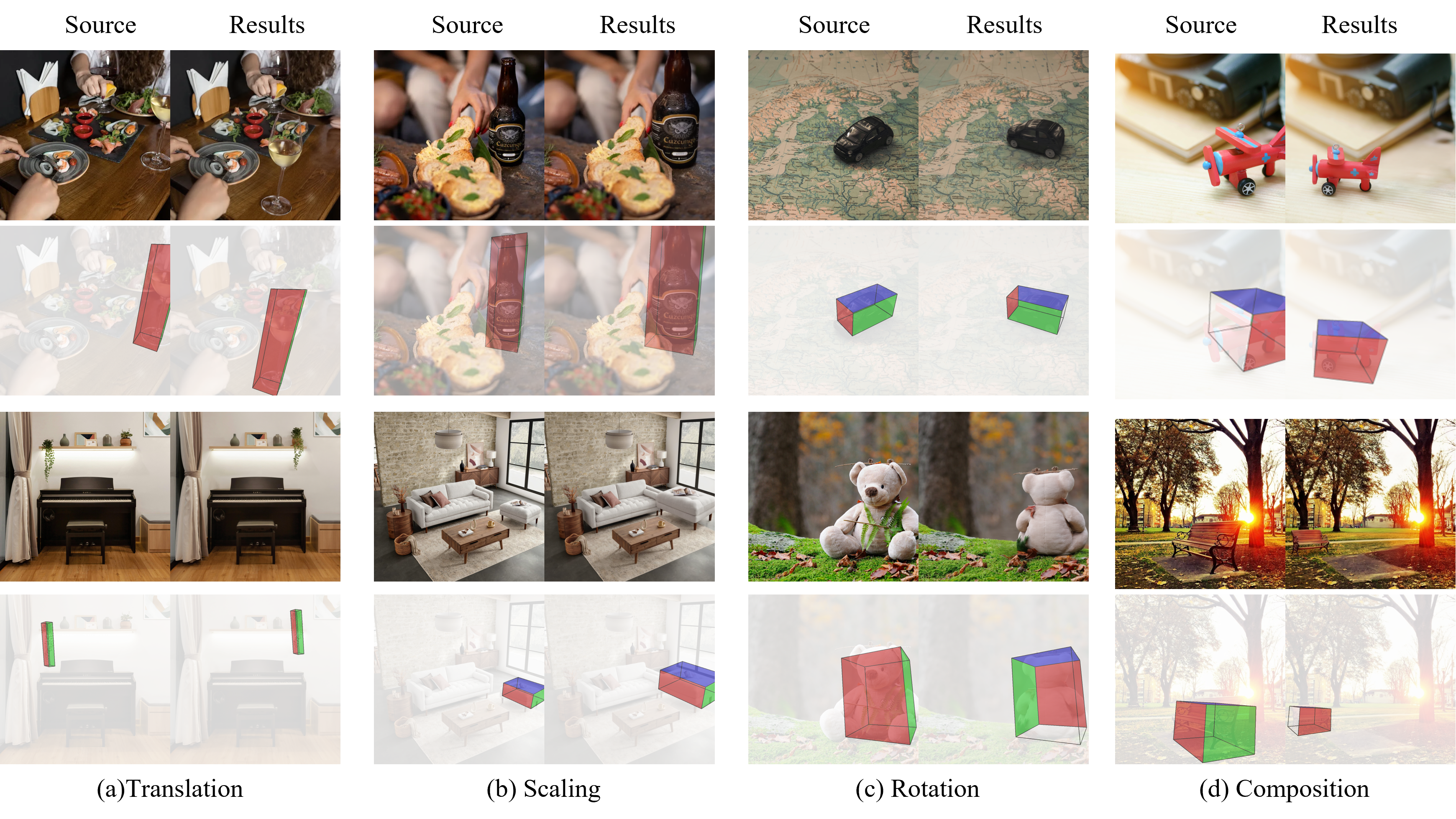}
    \caption{Our method takes as input a source image, its corresponding RGB 3D bounding box, and a user-specified transformed RGB 3D bounding box. The model then infers the geometric transformation between the two 3D boxes and faithfully transfers this transformation to the object in the source image.
Overall, BoxCtrl enables precise geometric editing, including translation, scaling, rotation, and their compositions within a unified framework, while preserving strong object identity and delivering high visual fidelity. Source images are from the ObjectMover Benchmark \cite{yu2025objectmover}.}
    \label{fig:teaser}
\end{teaserfigure}

\maketitle

\section{Introduction}

With the rapid development of instruction-based image editing models and multimodal large language models, using text prompts to control image editing has been well explored. 
However, textual descriptions are inherently ambiguous, making them more suitable for coarse-grained editing tasks. 
For geometric editing tasks that require precise spatial control, such as object translation, scaling, or rotation, text guidance alone is often insufficient for accurate manipulation, even with state-of-the-art models such as FLUX.1 Kontext~\cite{kontext}, or Qwen-Image-Edit~\cite{qwen-img}.

Compared to text conditions, visual conditions are more intuitive and allow more precise control. Therefore, numerous works have emerged exploring visual prompts for geometric image editing~\cite{sculpting, alzayer2025magic, 3d-fixup, freefine, chen2025blenderfusion, li2025blobctrl}. Based on the type of visual guidance, we divide these methods into 2D-guided methods and 3D-guided methods. 2D-guided methods~\cite{alzayer2025magic,freefine, li2025blobctrl, sajnani2025geodiffuser, diffusionhandles} usually adopt representations such as control points, bounding boxes, or blobs as the conditions. 2D guidance is more direct and user-friendly but inherently limited in editability. When dealing with large geometric transformations in 3D space (e.g., significant 3D rotations), these methods often fail to maintain structural consistency, resulting in distorted or implausible edits.

On the other hand, 3D-guided methods~\cite{sculpting,chen2025blenderfusion, 3d-fixup} use a 3D model as a proxy for editing. They typically require reconstructing a 3D model of the object, applying edits in 3D space, and then projecting the results back into 2D as the conditions for image generation. Although such methods can support diverse geometric edits, obtaining accurate 3D models is challenging.
Furthermore, direct manipulation of 3D assets demands a high level of expertise. This requirement for prior knowledge of 3D software presents a significant barrier to entry for non-professional users.

To address the limitations of existing methods, we propose 3D-aware visual prompting in a unified framework for precise and effective geometric image editing. We use 3D bounding boxes as our basic representation, as they naturally capture an object’s position and scale in 3D space. However, 3D bounding boxes are usually drawn as lines or painted in a single color, which contains position and scale information and fails to represent orientation. To address this, we propose the RGB 3D bounding box representation, where the three orthogonal faces of the box are painted in different colors, clearly indicating the object’s position, scale, and orientation as shown in Fig.~\ref {fig:teaser}. 

We adopt FLUX.1-Kontext-dev as our base model for its strong generation capability. However, existing image diffusion models lack precise controllability in 3D space. Furthermore, paired datasets with accurate annotations for geometric image editing are extremely scarce. To address this, we construct a large-scale synthetic dataset including diverse scenes, object categories, and geometric transformation parameters. For each training pair, the background and camera are fixed while the foreground object is randomly translated, rotated, or scaled. During rendering, the precise geometric editing parameters are recorded. The dataset can be easily scaled up to support a wide range of training scenarios and effectively fills the gap in existing resources for accurate supervised geometric editing training.

Leveraging our proposed visual prompting and curated paired large-scale synthetic dataset, we can easily perform Supervised Fine-Tuning (SFT) to equip the model with basic geometric editing capabilities. However, relying exclusively on SFT results in limited generalization because of the inherent domain gap between synthetic training data and real-world images.

To address this limitation, we leverage DiffusionNFT~\cite{zheng2025diffusionnft}, an online Reinforcement Learning (RL) approach, to explicitly optimize the model for superior geometric accuracy. This allows us to incorporate real-world training data without requiring paired ground truth. The success of RL hinges on the design of a robust reward function, which acts as the critical signal governing the final model performance. Effective geometric editing entails satisfying dual objectives: precise adherence to the given instructions and the preservation of visual fidelity. To balance these objectives, we design a comprehensive joint geometric reward function that efficiently steers the model toward superior editing accuracy. We leverage Grounding DINO~\cite{liu2024grounding} to detect the position and size of the edited object for evaluating translation and scaling precision, utilize OrientAnything V2~\cite{wang2026orient} to measure the rotation accuracy of the edited objects, and employ CLIP-I~\cite{radford2021learning} to ensure global visual fidelity. By optimizing the diffusion sampling with our well-designed reward function, we significantly enhance the model's adherence to geometric instructions, ensuring superior control accuracy while maintaining high image quality.

To validate our method, we construct both synthetic and real-world testing sets, along with comprehensive evaluation metrics to measure geometric editing accuracy. Experimental results demonstrate that our method exhibits strong generalization and achieves the best performance in image quality, object fidelity, and editing accuracy.

Our main contributions are summarized as follows:

\begin{itemize}[topsep=0pt, partopsep=0pt]
    \item We propose a unified framework guided by our novel 3D-aware visual prompting for precise geometric image editing. 
    \item We introduce a two-stage training strategy, first SFT and then RL, and design a comprehensive joint reward function. The reinforcement learning effectively improves the generalization of the SFT model on real data.
    \item We construct a well-annotated, large-scale synthetic dataset that enables sim-to-real geometric editing, covering object translation, scaling, rotation, and composite edits across a wide range of editing scenarios.
\end{itemize}

\label{sec:intro}
\section{Related work}
\subsection{Semantic Prompting}
3D-aware image editing approaches based on semantic prompting typically leverage transformation parameters as conditioning signals. For instance, Object3DIT~\cite{obj3dit} integrates encoded parameters with reference image embeddings, whereas Neural Assets~\cite{neural_assets} concatenates target 3D pose tokens with appearance features. However, such abstract semantic conditioning often constrains editing precision due to the inherent semantic-to-visual gap. To bridge this gap, our method introduces intuitive 3D-aware visual prompting: RGB 3D bounding boxes. By leveraging the latent visual priors of pretrained Diffusion Transformers (DiTs), this design facilitates significantly more precise and spatially-aware geometric control.

\subsection{Visual Prompting}
\paragraph {2D Guidance.}
Existing 2D-guided methods primarily rely on hand-crafted priors. For instance, Diffusion Handles~\cite{diffusionhandles} manipulates model activations based on estimated depth map transformations, while Magic Fixup~\cite{alzayer2025magic} and FreeFine~\cite{freefine} refine coarse crop-and-paste 2D edits using diffusion models. However, because they rely on 2D warping, these approaches are inherently limited in handling complex 3D transformations. In contrast, our approach conditions the diffusion process on source and target 3D bounding boxes, which explicitly encode the object’s position, scale, and orientation in 3D space.

\paragraph {3D Guidance.}
Methods such as Image Sculpting~\cite{sculpting}, BlenderFusion~\cite{chen2025blenderfusion}, and 3D Fixup~\cite{3d-fixup} leverage single-view reconstruction to lift and manipulate objects in 3D space. However, these pipelines are often labor-intensive and contingent upon reconstruction accuracy. To enhance efficiency and robustness, our approach performs geometric editing end-to-end directly within the latent space. By utilizing RGB 3D bounding boxes as dedicated spatial conditioning, our framework cleanly decouples spatial manipulation from appearance modeling, bypassing explicit 3D mesh recovery.

\subsection{3D-aware image generation}
While recent 3D-aware generation methods~\cite{min2025origen, koo2025videohandles, parihar2025zero, kumari2024customizing, parihar2025compass, omran2025controllable, bhat2024loosecontrol, eldesokey2024build} have made significant strides, they are often constrained by stringent input requirements or specific task formulations. Specifically, several approaches~\cite{kumari2024customizing, parihar2025compass} require multiple input views for personalization, whereas Omran et al.~\cite{omran2025controllable} target generative inpainting for new object placement. Most closely related to our work are LooseControl~\cite{bhat2024loosecontrol} and Build-A-Scene~\cite{eldesokey2024build}, which support the rotation editing of existing objects. However, LooseControl lacks precise orientation control due to its depth-only bounding boxes, and Build-A-Scene often introduces artifacts via warped latent guidance. In contrast, our approach enables accurate, photorealistic geometric editing from just a single image.

\section{Method}
\subsection{Overview}
\begin{figure*}[t]
    \centering
    \includegraphics[width=0.9\textwidth, height=0.6\textwidth, keepaspectratio]{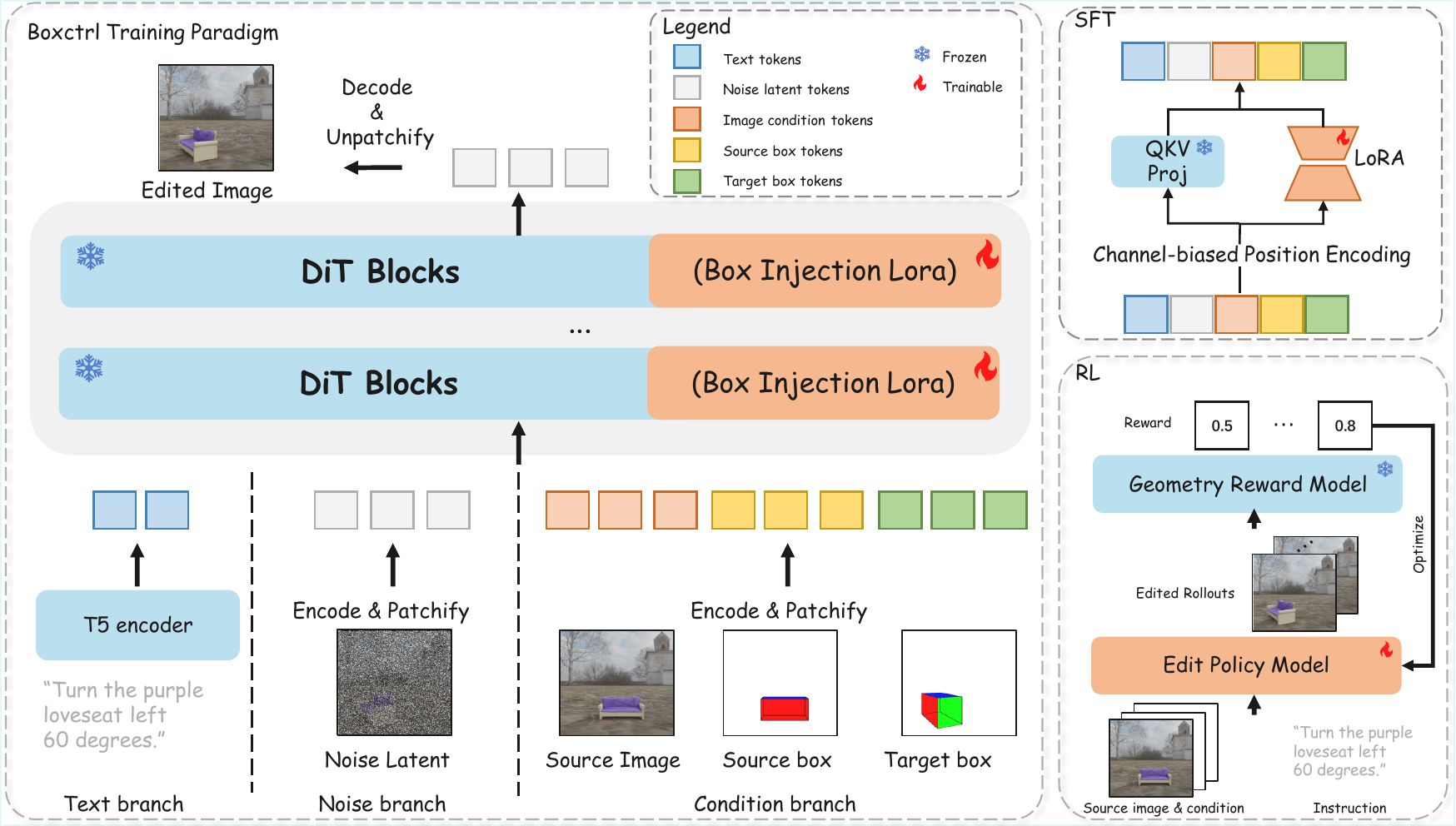}

    \caption{
    \textbf{Pipeline Overview.}
    Our method processes instructions, source images, and box pairs into conditional tokens for geometric image editing. 
    The model is trained via SFT followed by RL, utilizing a DiT-based policy optimized using GroundingDINO, OrientAnythingV2, and CLIP as geometry reward models.
    Key components include: (1) \textbf{Visual Prompting}: 2D-projected 3D boxes with RGB faces; (2) \textbf{Tokenization}: Concatenated sequence of text tokens $[B,L,D]$ and visual tokens $[B,4N,D]$; (3) \textbf{Box Injection LoRA}: Parallel LoRA integrated into QKV, output, and FFN layers; (4) \textbf{Channel-biased PE}: Modality-specific offsets to differentiate noise latent, image condition, source box and target box tokens.}
    \label{fig:pipeline}

\end{figure*}

The overall architecture of BoxCtrl is illustrated in Fig.~\ref{fig:pipeline}. Built upon a DiT backbone, our model utilizes pairs of source and target boxes as in-context visual prompting that enable consistent and precise control over translation, rotation, scaling, and composite edits within a single end-to-end framework.

In the subsequent sections, we first detail the task definition and the design of our 3D-aware visual prompting mechanism. Then we introduce the details of our SFT and RL training. Finally, we introduce our pipeline for SFT and RL training dataset curation, which offers a scalable solution for future large-scale training.

\paragraph {Task Definition.} Given a source image $I_s$, a pair of RGB 3D bounding boxes representing the source and target object poses, denoted as $B_s$ and $B_t$, and a text instruction $T$ describing the desired edit, our objective is to synthesize an edited image $I_t$ that reflects the transformation specified by $(B_s, B_t, T)$ while keeping all other elements in $I_s$ unchanged. The generation process can be formulated as:
\begin{equation}
    I_t = f_\theta(z_0 \mid I_s, B_s, B_t, T),
\end{equation}
where $z_0 \sim \mathcal{N}(\mathbf{0}, \mathbf{I})$ denotes a sampled noise latent, and $f_\theta$ represents the diffusion transformer.

\subsection{Visual Prompting}
\begin{figure}[t]
    \centering
    \includegraphics[width=\linewidth]{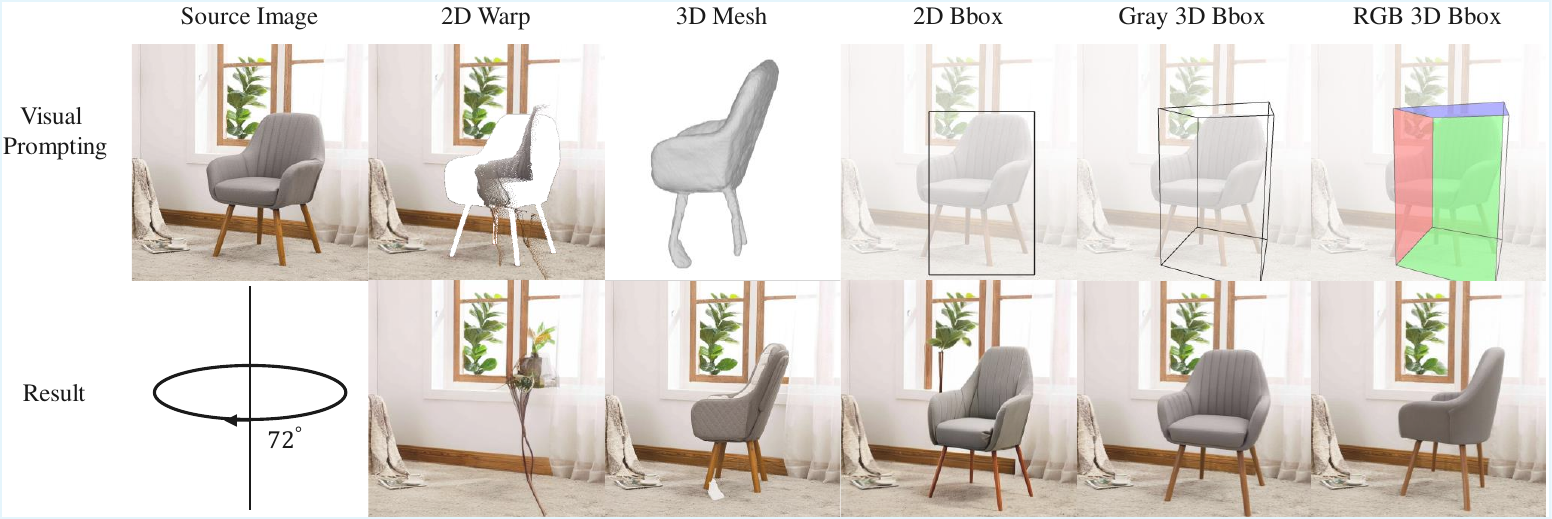}
    \caption{Comparison of visual prompting for geometric editing. We contrast 2D warping–based prompting and 3D mesh prompting with abstract visual prompting, including 2D bounding boxes, 3D gray bounding boxes, and our RGB 3D bounding boxes. The proposed RGB 3D bounding boxes produce notably more accurate geometric edits. Source Image and the 3D Mesh column (both prompting and result) are from Image Sculpting \cite{sculpting}.}

    \label{fig:single}
\end{figure}

\paragraph {Orientation-aware 3D Bounding Box.} Visual prompting provides intuitive and effective guidance for geometric editing. Existing works can be divided into two categories: detailed visual prompting
\cite{sculpting, freefine, alzayer2025magic, 3d-fixup, chen2025blenderfusion, i2v3d} and abstract visual prompting \cite{li2025blobctrl, chen2024anydoor}, as shown in Fig.~\ref{fig:single}. Detailed visual prompting obtains a reference image of the object after geometric transformation through 2D warping \cite{freefine, alzayer2025magic}, or 3D mesh reconstruction and rendering \cite{sculpting, 3d-fixup, chen2025blenderfusion, i2v3d}, to guide the generation of edited images. However, such pixel-level guidance is fragile: 2D warping often distorts object geometry under large rotations, while 3D mesh–based prompting suffers from reconstruction errors. Abstract visual prompting is more robust, but existing forms are limited in representing 3D transformations. For example, a 2D bounding box\cite{chen2024anydoor} cannot express 3D transformations, while a 3D gray bounding box can describe object position and scale but still has ambiguity in representing orientation. 

Our goal is to integrate position, scale, and orientation into a single visual prompting without introducing redundant parameters, while ensuring that the prompting remains sufficiently discriminative across different editing configurations. To achieve this, we encode object orientation by colorizing the box surfaces. When the model observes the rotation of surfaces with consistent colors, it can infer the intended orientation change and apply the corresponding rotation to the object in the source image. This enables the only successful case of chair rotation.

\label{sec:visual_prompt}

Building upon our proposed visual prompting mechanism, we implement a two-stage training framework. We first conduct SFT using our paired synthetic dataset, followed by an online RL stage trained on unpaired real-world data.

\subsection{Supervised Fine-tuning}
We perform SFT to adapt the model for geometric editing tasks. To achieve parameter-efficient fine-tuning, we inject Low-Rank Adaptation (LoRA)~\cite{lora} weights into the attention projection layers and linear layers within the DiT blocks. During training, the model accepts a tuple input comprising the source image, source bounding box, target bounding box, and the text editing instruction. The network is optimized using the flow matching loss~\cite{flowmatching}.

\subsection{Reinforcement Learning}
To enhance the model's generalization to real-world data, we employ an RL stage adapted from the DiffusionNFT~\cite{zheng2025diffusionnft}. The training adheres to an iterative rollout-update cycle: the policy model generates rollouts, receives feedback from the reward model, and updates its parameters to maximize the expected reward.

\paragraph{DiffusionNFT Preliminary.}
DiffusionNFT is a post-training reinforcement learning framework designed to optimize the generation process of diffusion models. 
Intuitively, the diffusion process "steers" a trajectory from noise to a clean image, and DiffusionNFT optimizes this steering velocity by pulling the model toward high-reward trajectories and pushing it away from low-reward ones. 
During fine-tuning, the active policy $\theta$ (e.g., a DiT) first generates context-conditioned samples $\{\boldsymbol{x}_0^k\}_{k=1}^K$ to receive normalized rewards $r \in [0, 1]$. By adding forward noise to $\boldsymbol{x}_0$, the intermediate state $\boldsymbol{x}_t$ is constructed. At each state $\boldsymbol{x}_t$, DiffusionNFT optimizes the following weighted flow matching loss to encourage $\boldsymbol{v}_{\theta}$ to follow high-reward paths while penalizing low-reward ones:
\begin{equation}
\mathbb{E}_{\boldsymbol{c}, \pi^{\text{old}}(\boldsymbol{x}_0|\boldsymbol{c}), t} \left[ r\|\boldsymbol{v}^+_\theta(\boldsymbol{x}_t, \boldsymbol{c}, t) - \boldsymbol{v}\|_2^2 + (1 - r)\|\boldsymbol{v}^-_\theta(\boldsymbol{x}_t, \boldsymbol{c}, t) - \boldsymbol{v}\|_2^2 \right],
\end{equation}
\begin{align*}
\boldsymbol{v}^+_\theta(\boldsymbol{x}_t, \boldsymbol{c}, t) &:= (1 - \beta)\boldsymbol{v}^{\text{old}}(\boldsymbol{x}_t, \boldsymbol{c}, t) + \beta\boldsymbol{v}_\theta(\boldsymbol{x}_t, \boldsymbol{c}, t),\\
\boldsymbol{v}^-_\theta(\boldsymbol{x}_t, \boldsymbol{c}, t) &:= (1 + \beta)\boldsymbol{v}^{\text{old}}(\boldsymbol{x}_t, \boldsymbol{c}, t) - \beta\boldsymbol{v}_\theta(\boldsymbol{x}_t, \boldsymbol{c}, t),
\end{align*}
where $\boldsymbol{v}^{\text{old}}$ denotes the old policy, $\boldsymbol{v}_{\theta}^{\pm}$ are the implicit positive and negative policies, $\boldsymbol{v}$ is the target velocity pointing to the generated sample $\boldsymbol{x}_0$, and $\beta$ controls the reinforcement strength.

Our reward function is meticulously designed to balance geometric precision with visual quality, composed of three key components: (1) spatial alignment for translation and scaling, (2) orientation alignment for rotation, and (3) image fidelity for preserving visual details.

\paragraph {Translation and Scaling Reward.}
The translation and scaling reward is defined using the Intersection over Union (IoU), which simultaneously captures both the position and size of the edited object.
We then detect the bounding box of the object in the edited image using GroundingDINO~\cite{liu2024grounding}, denoted as $B_{\text{pred}}$. 
The reward is computed as the IoU between the predicted and target bounding boxes:
\begin{equation}
r_{\text{trans\_scale}}
= \text{IoU}(B_{\text{pred}}, B_{\text{tgt}})
= \frac{\text{Area}(B_{\text{pred}} \cap B_{\text{tgt}})}{\text{Area}(B_{\text{pred}} \cup B_{\text{tgt}})}.
\end{equation}

\paragraph {Rotation Reward.}
The rotation reward $r_{\text{rot}}$ is designed to enforce rotation angular precision.
Let $\theta_{\text{tgt}}$ denote the target rotation angle specified by the editing instruction.
We employ OrientAnything-v2~\cite{wang2026orient} to estimate the predicted rotation angle of the object in the input image and in the edited image, denoted as $\theta_{\text{pred}} \in [0^\circ, 360^\circ]$.
The reward is formulated as
\begin{equation}
r_{\text{rot}} = 0.5 + 0.5 \cos\!\left(\theta_{\text{pred}} - \theta_{\text{tgt}} \right).
\end{equation}

\paragraph {Joint Reward Formulation.}
To ensure both geometric accuracy and visual fidelity, the joint reward is computed as a weighted sum of specific objectives:
\begin{equation}
\begin{split}
    r(I_{\text{pred}}, I_{\text{src}}, T) =\; & \lambda_{\text{sim}} r_{\text{sim}}(I_{\text{pred}}, I_{\text{src}}) \\
    & + \lambda_{\text{ts}} r_{\text{trans\_scale}}(I_{\text{pred}}) \\
    & + \lambda_{\text{rot}} r_{\text{rot}}(I_{\text{pred}}),
\end{split}
\label{eq:joint_reward}
\end{equation}
where $I_{\text{src}}$ is the input image, $I_{\text{pred}}$ is the edited image, $\lambda_{\text{sim}}$, $\lambda_{\text{ts}}$, and $\lambda_{\text{rot}}$ are weighting coefficients for each reward component. The term $r_{\text{sim}}$ represents the image similarity reward, calculated via CLIP image similarity ~\cite{radford2021learning}.

To ensure the stability of the RL optimization, we introduce a zero-reward fallback strategy to address detection failures. Specifically, generated samples that fail to meet predefined confidence thresholds are classified as invalid and strictly assigned a zero reward to penalize the policy. We empirically set the thresholds to 0.25 for GroundingDINO and 0.5 for OrientAnythingV2, which constrains the corresponding failure rates to approximately $1\%$ and $8\%$, respectively.

\subsection{Dataset Construction}

We leverage a hybrid dataset comprising both synthetic and real-world data for the SFT and RL stages, respectively. A key advantage of our approach is that the data acquisition pipelines for both stages are highly scalable.
\label{sec:kubric_tool}

\paragraph{SFT Dataset Curation}
As shown in Tab.~\ref{tab:dataset_comparison}, there are very few geometric datasets containing fixed camera and scene settings, accurate 3D transformation parameters $P$, precise 3D bounding box annotations, and detailed editing instructions.
Video-based datasets are often unsuitable because they inherently couple camera motion with object motion, 
making it difficult to obtain high-quality samples where only the object’s spatial position, scale, or orientation changes 
while the background and camera remain fixed. 

To address this issue, we construct a dataset of 23,976 samples about 235 objects with precise transformation annotations using the public Kubric engine~\cite{kubric}, 
built on top of Blender and the 3D scanned object dataset GSO~\cite{gso}. 
Specifically, we filter out 235 visually clean and category-diverse objects from GSO. 
For each object, we fix the camera viewpoint, randomly choose an HDRI environment as background, 
and place the object in the scene. 
We then apply random transformations, including translation (range $[-0.5, 0.5]$), 
scaling (range $[0.3, 3.0]$), and rotation along the $y$-axis (range $[-180, 180]$), 
while adjusting the object to prevent unnatural object-ground penetration. 
Each transformation yields paired images together with corresponding bounding box images.

To ensure diversity and efficiency, for each object we uniformly sample rotations every $30^\circ$, and form ordered pairs as source and target images. 
During dataset composition, we control the number of samples per rotation angle to maintain a uniform distribution. 
Finally, we employ Gemini-2.5-Flash~\cite{gemini2.5} to generate textual captions for each object, resulting in a comprehensive paired dataset that includes the source image, source box, target image, target box, and corresponding edit instruction prompts.

\begin{table}[htbp]
    \centering
    \small
    \setlength{\tabcolsep}{1 pt}
    \caption{Comparison of geometric editing datasets. Our dataset is the most comprehensive geometric editing dataset to date, containing high-quality pairs with accurate 3D transformation parameters $P$, precise 3D bounding box annotations, and detailed editing instructions.}
    \label{tab:dataset_comparison}
    \begin{tabular}{p{4cm}|c|c|c} 
        \toprule
        Dataset & Accurate $P$ & 3D Bboxes & Edit instruction \\
        \midrule
        Obj3DIT ~\cite{obj3dit} & \ding{51} &  \ding{55} & \ding{51} \\
        Magic Fixup ~\cite{alzayer2025magic} &  \ding{55} &  \ding{55} &  \ding{55} \\
        3D Fixup  ~\cite{3d-fixup} &  \ding{55} &  \ding{55} &  \ding{55} \\
        Nerual Assets ~\cite{neural_assets} & \ding{51} & \ding{51} &  \ding{55} \\
        Blenderfusion ~\cite{chen2025blenderfusion} & \ding{51} & \ding{51} &  \ding{55}\\
        \midrule
        Ours  & \ding{51} & \ding{51} & \ding{51}\\
        \bottomrule
    \end{tabular}
\end{table}

\paragraph{RL Dataset Curation}
Our RL dataset comprises 1,164 unpaired editing samples. The base images are sourced from PIE-Bench \cite{ju2023direct} and Subjects200k \cite{tan2025ominicontrol}, spanning diverse categories: humans (20$\%$), animals (40$\%$), and scenes (40$\%$).
We procedurally generate editing instructions encompassing translation, rotation, scaling, and composite transformations. Specifically, we utilize DetAny3D~\cite{zhang2025detect} to estimate the initial 3D bounding boxes of objects within the scene. By applying the instruction-specified transformations to these initial poses, we derive the target 3D bounding boxes. The two bounding boxes serve as the geometric conditioning for the model. Crucially, this pipeline obviates the need for ground-truth edited images, thereby enabling the effective utilization of unpaired real-world data during RL training.

\section{Experiments}
\subsection{Implementation Details}

Initialized with FLUX.1-Kontext-dev~\cite{kontext}, our model is trained on $512 \times 512$ images. We conduct SFT for 26K steps with LoRA rank 128 and $lr=1\times10^{-4}$, followed by RL for 60 epochs with LoRA rank 32 and $lr=3\times10^{-4}$. During inference, source boxes are detected via DetAnything3D~\cite{zhang2025detect}, while target boxes are synthesized using random translation, scaling, rotation, or combinations thereof. For synthetic evaluation, we curate 100 unseen samples using the tools in Section~\ref{sec:kubric_tool}.  
For real-world validation, we evaluate approximately 100 samples per editing task from public datasets ~\cite{ju2023direct, tan2025ominicontrol, yu2025objectmover}. 

\paragraph{Baselines.} We compare our approach with state-of-the-art open-source models across three paradigms: instruction-based editing (FLUX.1-Kontext-dev~\cite{kontext}, Qwen-Image-Edit~\cite{qwen-img}), visual prompting-based editing (Magic Fixup~\cite{alzayer2025magic}, FreeFine~\cite{freefine}), and 3D-aware generation (LooseControl~\cite{bhat2024loosecontrol}, Build-A-Scene~\cite{eldesokey2024build}).

\subsection{Evaluation Metrics}
For evaluation on the synthetic dataset with ground-truth annotations and images, we assess the geometric editing performance from the following three aspects:

\begin{itemize}
    \item \textbf{Image Quality.} 
    We use standard image quality metrics, including PSNR, SSIM~\cite{ssim}, and LPIPS~\cite{lpips} to evaluate the overall image quality.

    \item \textbf{Identity Preservation.} 
    We employ DINO-I~\cite{dino} and CLIP-I ~\cite{radford2021learning} to measure the high-level semantic similarity between the edited objects and ground-truth, quantifying how well the object identity is preserved under geometric transformations.

    \item \textbf{Geometric Accuracy.} 
    We utilize Grounding-SAM \cite{ground-sam} to extract 2D masks from both the edited and ground-truth images, and compute the IoU between them to capture spatial alignment and measure the rotation accuracy (RotAcc) of the generated orientations relative to the targets using OrientAnything-V2~\cite{wang2026orient}. 
\end{itemize}

For evaluation on real-world data, we evaluate the geometric accuracy, detect the bounding box IoU via Grounding-DINO \cite{liu2024grounding} to assess translation and scaling performance, and compute the rotation accuracy between the generated and target orientations using OrientAnything-V2 \cite{wang2026orient}, and assess visual fidelity using CLIP-I and DINO-I.
We also conducted a human evaluation, with details provided in the Supplementary Material S4.

\subsection{Quantitative Evaluation}
As shown in Tab.~\ref{tab:compare}, FLUX-Kontext yields high similarity scores by trivially reconstructing the input, largely ignoring geometric instructions. Qwen-Image-Edit suffers from inaccurate spatial mapping due to text ambiguity. Although FreeFine and Magic Fixup demonstrate competitive performance in translation and scaling, they struggle significantly with complex rotation and composite tasks. Specifically, LooseControl lacks precise orientation control due to its depth-aware-only boxes. Meanwhile, Build-A-Scene introduces noticeable artifacts via warped latent guidance. Furthermore, both methods require inversion to process real images, causing an inherent editability-fidelity trade-off. In contrast, our method delivers precise translation, scaling, rotation, and composition editing in a unified framework, benefiting from our proposed 3D-aware visual prompting to explicitly represent geometric intentions.

\begin{table*}[t]
\centering
\caption{\textbf{Comprehensive comparison on our benchmark.} We evaluate our method against state-of-the-art image editing methods across four tasks. $\uparrow$ indicates higher is better. \textbf{Bold} and \underline{underline} highlight the best and second-best results.}
\label{tab:comparison}
\resizebox{\textwidth}{!}{
\begin{tabular}{l|ccc|ccc|ccc|cccc}
\hline
\multirow{2}{*}{Method} & \multicolumn{3}{c|}{Translation} & \multicolumn{3}{c|}{Scaling} & \multicolumn{3}{c|}{Rotation} & \multicolumn{4}{c}{Composite} \\
\cline{2-14}
 & IoU$\uparrow$ & CLIP-I $\uparrow$ & DINO-I $\uparrow$
 & IoU$\uparrow$ & CLIP-I $\uparrow$  & DINO-I $\uparrow$
 & Rot Acc $\uparrow$ & CLIP-I $\uparrow$  & DINO-I $\uparrow$
 & IoU$\uparrow$ & Rot Acc $\uparrow$ & CLIP-I$\uparrow$ & DINO-I $\uparrow$ \\
\hline
FLUX-Kontext    
&0.413 	&\textbf{0.977} &\textbf{0.975} &0.513 &\textbf{0.978}  &\textbf{0.974}	&0.515 	&\textbf{0.980}  &\textbf{0.974} &0.456 	&0.554 	&\textbf{0.968} &\textbf{0.967}
\\
Qwen-Image-Edit 
&0.339 	&0.948 &0.948 &0.492&0.937 &0.897
&\underline{0.695} 	&0.910	 &0.857
&0.352 &0.634&0.931 &0.927
 \\
Magic Fixup    
&0.882 	&0.940 	 &0.936	&0.810 	&0.935 	&0.909	&0.511 	&0.844 	&0.650	&0.509 	&0.626 	&0.822 &0.616
\\
FreeFine 
&\textbf{0.912} 	&0.944 	&0.944	&\textbf{0.820} 	&0.938 &0.907		&0.585 	&0.830 	&0.591	&0.458 	&0.572 	&0.812 &0.571
\\
LooseControl &0.407 &0.945 	&0.931 	&0.486 	&0.951 	&0.935 	&0.505 	&\underline{0.956} 	&\underline{0.942} 	&0.444 	&0.549 	&\underline{0.945} 	& \underline{0.928} \\
Build-A-Scene &0.821 	&0.867 	&0.822 	&0.774 	&0.846 	&0.727 	&0.614 	&0.755 	&0.471 	&\underline{0.534} 	&\underline{0.641} 	&0.731 	&0.448 \\

\hline
\textbf{Ours} 
&\underline{0.889} 	&\underline{0.957} &	\underline{0.970}	&\underline{0.816} 	&\underline{0.957} &\underline{0.951} &\textbf{0.899} &0.901 &0.815	&\textbf{0.682} &\textbf{0.841} &0.894 &0.814
  \\
\hline
\end{tabular}
}
\label{tab:compare}
\end{table*}

\subsection{Qualitative Evaluation}
\begin{figure*}[htbp]
    \centering
    \includegraphics[width=0.95\textwidth, height=\textwidth, keepaspectratio]{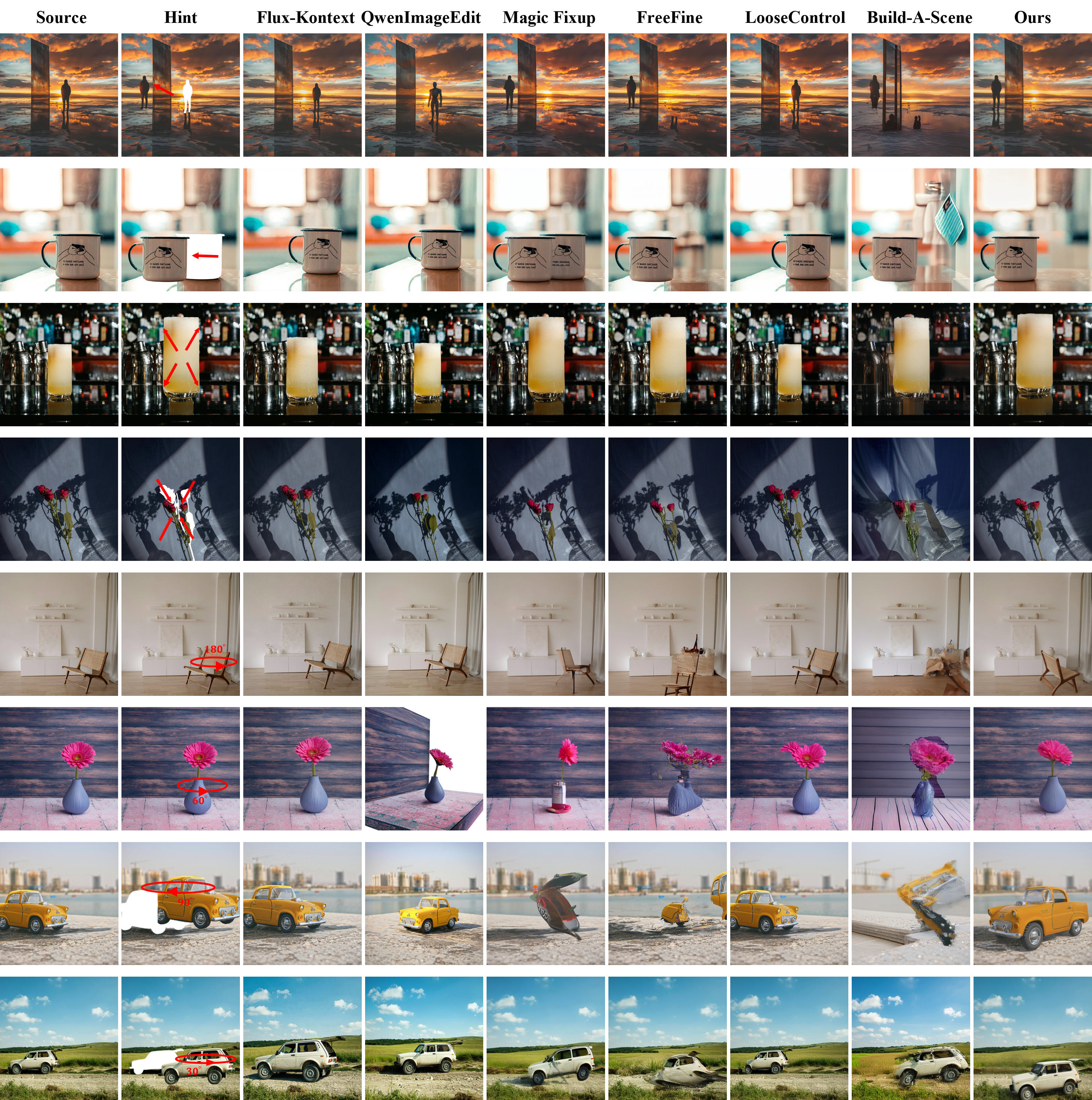}
    \caption{
        \textbf{Qualitative comparison with state-of-the-art image editing methods.} We evaluate our approach against instruction-based baselines (FLUX-Kontext, Qwen-Image-Edit), visual prompting-based methods (FreeFine, Magic Fixup), and 3D-aware image generation methods (LooseControl, Build-A-Scene). The results span four categories: translation (Rows 1-2), scaling (Rows 3-4), rotation (Rows 5-6), and composite editing (Rows 7-8). Compared to baselines, our method demonstrates superior geometric precision and visual fidelity, particularly in complex scenarios. Source images are from the ObjectMover Benchmark \cite{yu2025objectmover}.
    }
    \label{fig:comparison}
\end{figure*}

Fig.~\ref{fig:comparison} presents qualitative comparisons across four geometric editing tasks. Our method demonstrates superior performance across all editing tasks. 
Specifically, for translation, FLUX-Kontext, Qwen-Image-Edit, and LooseControl show minimal object movement. Magic Fixup fails to produce harmonious edits because it improperly injects features from the reference image. FreeFine produces blurry images due to the separate processing of foreground and background, and Build-A-Scene introduces artifacts. In contrast, our method achieves precise translation while maintaining natural lighting and shadows, resulting in visually realistic outputs.
In the scaling task, FLUX-Kontext, Qwen-Image-Edit, and LooseControl exhibit limited sensitivity to scale changes, resulting in insufficient magnification. While Magic Fixup, FreeFine, and Build-A-Scene can adhere to coarse edits, they often produce unrealistic backgrounds. In contrast, our method achieves precise scaling while maintaining the most natural object placement.
For rotation, FLUX-Kontext and LooseControl often under-rotate objects, while Qwen-Image-Edit alters the surrounding scene, leading to spatial inconsistency. Magic Fixup, FreeFine, and Build-A-Scene introduce noticeable distortions and artifacts from warping. Our approach performs accurate rotations without unrealistic deformations. 
Finally, in compositional editing, only our BoxCtrl successfully controls position, scale, and orientation at the same time. FLUX-Kontext, Qwen-Image-Edit, and LooseControl struggle with instructions, while Magic Fixup, FreeFine, and Build-A-Scene fail to preserve object structure because of the limited capabilities of warping.

\subsection{Ablation Study}

\paragraph{Ablation on Condition Type.}
We ablate three conditions: (1) text-only, (2) RGB 3D bboxes only, and (3) RGB 3D bboxes + text (ours). As shown in Tab. ~\ref{tab:condition_type} and Fig.~\ref{fig:condition_type}, text-only inputs lack precise spatial control due to linguistic ambiguity, whereas RGB 3D bboxes enable reliable manipulation. With and without text yield similar performance, proving our RGB 3D bboxes visual prompting dominates generation.
\begin{table}[htbp]
  \centering
  \caption{Ablation of different condition types on the synthetic dataset.}

  \label{tab:ablation}
  \setlength{\tabcolsep}{2pt} 
  \small 
  \begin{tabular}{@{}lccccccc@{}}
    \toprule
    Setting & PSNR$\uparrow$ & SSIM$\uparrow$ & LPIPS$\downarrow$ & IoU$\uparrow$ & RotAcc$\uparrow$ & CLIP-I$\uparrow$ & DINO-I$\uparrow$ \\
    \midrule
    text &22.588 	&0.901	&0.148	&0.30	&0.641	&0.913 &0.924\\
    bbox & \textbf{26.379} & \textbf{0.921}	&\textbf{0.087}	&\textbf{0.88}	&0.861 &0.941	&\textbf{0.956}\\
    bbox + text &26.317	&0.920	&\textbf{0.087}	&\textbf{0.88}	&\textbf{0.879} &\textbf{0.944}	&\textbf{0.956} \\
    \bottomrule
  \end{tabular}
  \label{tab:condition_type}
\end{table}

\paragraph{Ablation on Visual Prompting.} We conducted an ablation study on the design of the visual prompting by comparing three configurations: (1) gray-scale source and target bboxes, (2) target-only RGB bboxes, and (3) RGB source and target bboxes (our final design). As shown in Tab.~\ref{tab:visual_abl} and Fig.~\ref{fig:ablation_box}, compared to gray-scale bboxes, RGB bboxes help the model capture face correspondences and distinguish orientations more effectively, thereby reducing ambiguity during editing. Additionally, providing both source and target bboxes as in-context visual examples significantly improves editing accuracy compared to using only target bboxes. This enhancement arises from the model’s ability to better learn relative transformations, enabling it to generalize bbox-based geometric changes to real images.

\begin{table}[t]
  \centering
  \caption{Ablation on visual prompting on the synthetic dataset.}

  \label{tab:ablation}
  \setlength{\tabcolsep}{2pt} 
  \small 
  \begin{tabular}{@{}lccccccc@{}}
    \toprule
    Setting & PSNR$\uparrow$ & SSIM$\uparrow$ & LPIPS$\downarrow$ & IoU$\uparrow$ & RotAcc$\uparrow$ & CLIP-I$\uparrow$ & DINO-I$\uparrow$ \\
    \midrule
    dual gray bboxes  &25.139	&0.914	&0.103 &0.80	&0.624	&0.928 & 0.947 \\
    single RGB bbox &25.859	&0.917	&0.097 &0.84	&0.752	&0.935 & 0.952 \\
    dual RGB bboxes &\textbf{26.317}	&\textbf{0.920}	&\textbf{0.087}	&\textbf{0.88}	&\textbf{0.879} &\textbf{0.944}	&\textbf{0.956}\\
    \bottomrule
  \end{tabular}
  \label{tab:visual_abl}
\end{table}

\paragraph{Ablation on RL} To validate the effectiveness of RL, we conduct a comprehensive ablation study on both synthetic and real datasets. As shown in Tab.~\ref{tab:rl_ablation}, Fig.~\ref{fig:sft_rl_syn}, and Fig.~\ref{fig:reward_function}, SFT alone is sufficient for the synthetic dataset; on the real dataset, the geometric precision of the SFT-only model degrades significantly. Here, the integration of RL effectively bridges the sim-to-real gap, yielding superior geometric editing precision.

\begin{table}[htbp]
  \centering
  \caption{Ablation of RL on the synthetic and real datasets.}

  \label{tab:combined_ablation}
  \setlength{\tabcolsep}{3pt} 
  \small 
  \begin{tabular}{@{}lcccccccc@{}}
    \toprule
    \multicolumn{8}{c}{\textbf{Synthetic Dataset}} \\
    \midrule
    Setting & PSNR$\uparrow$ & SSIM$\uparrow$ & LPIPS$\downarrow$ & IoU$\uparrow$ & RotAcc$\uparrow$ & CLIP-I$\uparrow$ & DINO-I$\uparrow$ \\
    \midrule
    SFT     & \textbf{26.317} & \textbf{0.920} & \textbf{0.087} & \textbf{0.88} & \textbf{0.879} & \textbf{0.944} & \textbf{0.956} \\
    SFT+RL  & 25.881 & 0.918 & 0.099 & 0.87 & 0.850 & 0.942 & 0.952 \\
    \midrule
    \multicolumn{8}{c}{\textbf{Real Dataset}} \\
    \midrule
    Setting & \multicolumn{2}{c}{IoU$\uparrow$} & \multicolumn{2}{c}{RotAcc$\uparrow$} & \multicolumn{2}{c}{CLIP-I$\uparrow$} & DINO-I$\uparrow$ \\
    \midrule
    SFT     & \multicolumn{2}{c}{0.618} & \multicolumn{2}{c}{0.752} & \multicolumn{2}{c}{\textbf{0.924}} & \textbf{0.866} \\
    SFT+RL  & \multicolumn{2}{c}{\textbf{0.682}} & \multicolumn{2}{c}{\textbf{0.841}} & \multicolumn{2}{c}{0.894} & 0.814 \\
    \bottomrule
  \end{tabular}
  \label{tab:rl_ablation}
\end{table}

\paragraph{Ablation on Reward Function Component.} To justify each component in our reward design, we conduct an ablation study across five settings: (1) SFT-only, (2) w/o rotation reward, (3) w/o translation/scaling reward, (4) w/o image similarity reward, and (5) the full joint reward. Quantitative (Tab.~\ref{tab:reward_function}) and qualitative (Fig.~\ref{fig:reward_function}) results confirm that RL-based fine-tuning consistently surpasses the SFT baseline, particularly in real-world generalization. Ablating individual geometric rewards leads to suboptimal performance in corresponding geometric metrics, whereas the joint strategy holistically boosts accuracy across all geometric dimensions. Furthermore, the image similarity reward proves essential for maintaining high visual fidelity and preserving fine-grained details.

\begin{table}[htbp]
  \centering
  \caption{Ablation of reward function component on the real dataset.}
  \label{tab:ablation}
  \begin{tabular}{@{}lcccc@{}}
    \toprule
    Setting & IoU $\uparrow$ & Rot Acc $\uparrow$ & CLIP-I  $\uparrow$ & DINO-I  $\uparrow$\\
    \midrule
    SFT  &0.618 	&0.752 	&\textbf{0.924} & 0.866  \\
    w/o $r_{\text{rot}}$ &\textbf{ 0.803} 	&0.605 	&\textbf{0.924} &\textbf{0.882}   \\
    w/o $r_{\text{trans\_scale}}$ & 0.609 	&0.881 	&0.884 &0.778  \\
    w/o $r_{\text{sim}}$ &0.696 &\textbf{0.917} &0.879 &0.752  \\
    Ours &0.682 &0.841 &0.894 &0.814  \\
    \bottomrule
  \end{tabular}
  \label{tab:reward_function}
\end{table}

\subsection{Challenging scenes}
\begin{figure}[htbp] 
    \centering
    \captionsetup[subfigure]{skip=3pt} 
    
    \begin{subfigure}[b]{0.48\linewidth}
        \centering
        \begin{tabular}{@{}cc@{}}
            \textbf{Source} & \textbf{Edit} \\
            \includegraphics[width=0.48\linewidth]{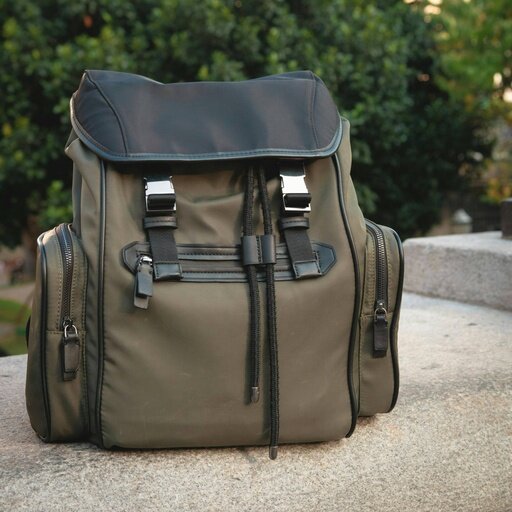} &
            \includegraphics[width=0.48\linewidth]{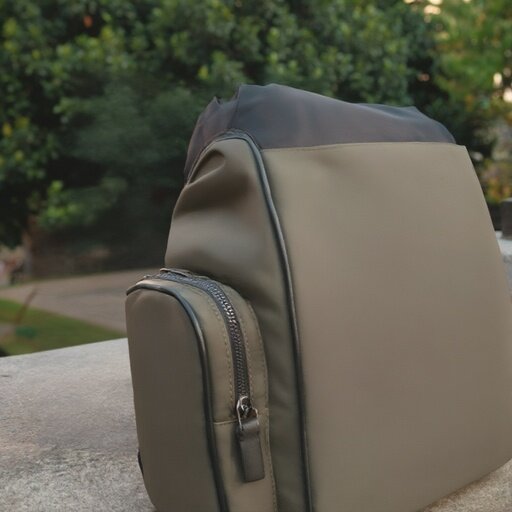}
        \end{tabular}
        \caption{Large geometric transformations}
        \label{fig:large_geom}
    \end{subfigure}
    \hfill
    \begin{subfigure}[b]{0.48\linewidth}
        \centering
        \begin{tabular}{@{}cc@{}}
            \textbf{Source} & \textbf{Edit} \\
            \includegraphics[width=0.48\linewidth]{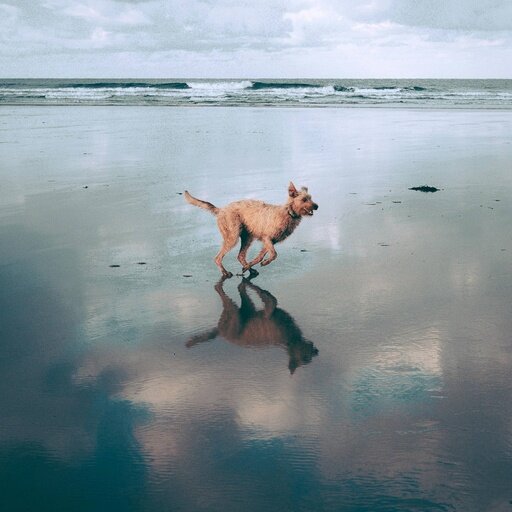} &
            \includegraphics[width=0.48\linewidth]{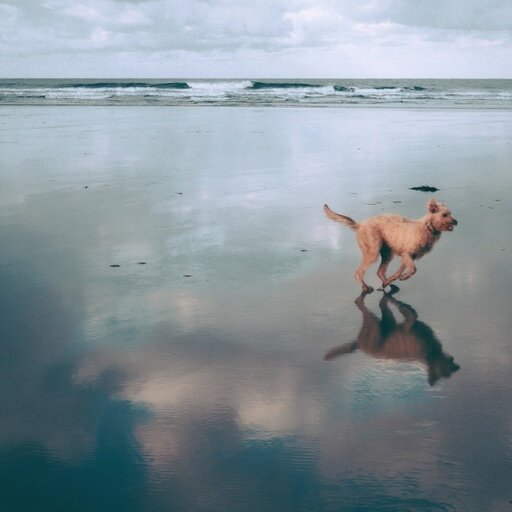}
        \end{tabular}
        \caption{Reflective surfaces}
        \label{fig:reflective}
    \end{subfigure}

    \begin{subfigure}[b]{0.48\linewidth}
        \centering
        \begin{tabular}{@{}cc@{}}
            \textbf{Source} & \textbf{Edit} \\
            \includegraphics[width=0.48\linewidth]{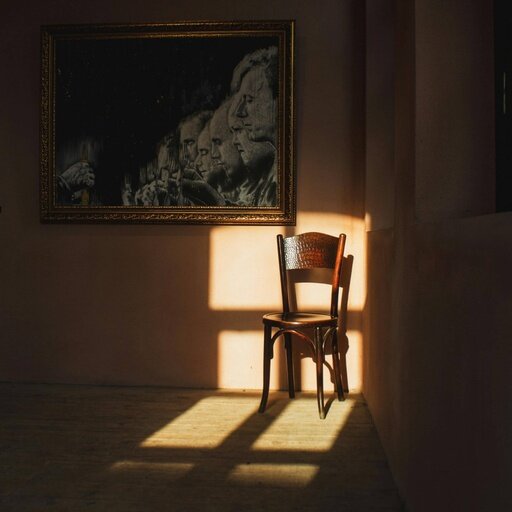} &
            \includegraphics[width=0.48\linewidth]{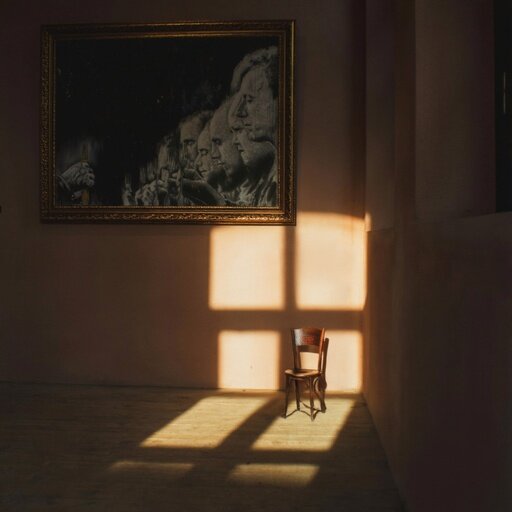}
        \end{tabular}
        \caption{Complex illumination}
        \label{fig:illumination}
    \end{subfigure}
    \hfill
    \begin{subfigure}[b]{0.48\linewidth}
        \centering
        \begin{tabular}{@{}cc@{}}
            \textbf{Source} & \textbf{Edit} \\
            \includegraphics[width=0.48\linewidth]{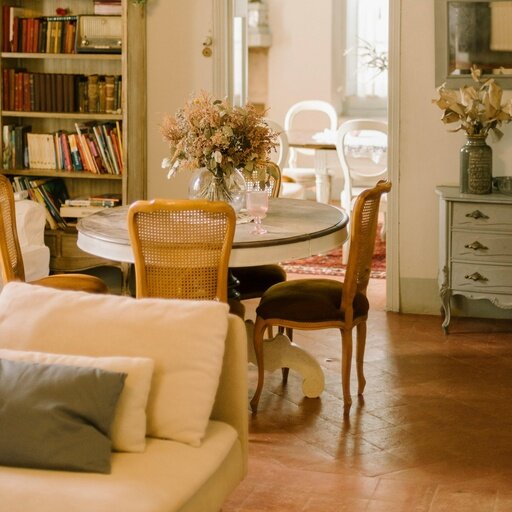} &
            \includegraphics[width=0.48\linewidth]{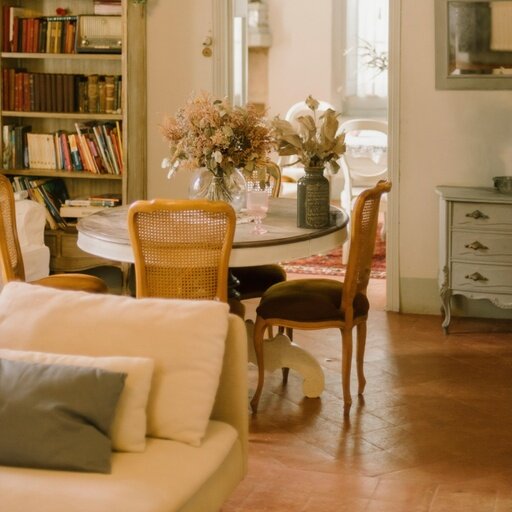}
        \end{tabular}
        \caption{Crowded scene structures}
        \label{fig:crowded}
    \end{subfigure}

    \begin{subfigure}[b]{\linewidth}
        \centering
        \begin{tabular}{@{}cccc@{}}
            \textbf{Source} & \textbf{Edit} & \textbf{Source} & \textbf{Edit} \\
            \includegraphics[width=0.23\linewidth]{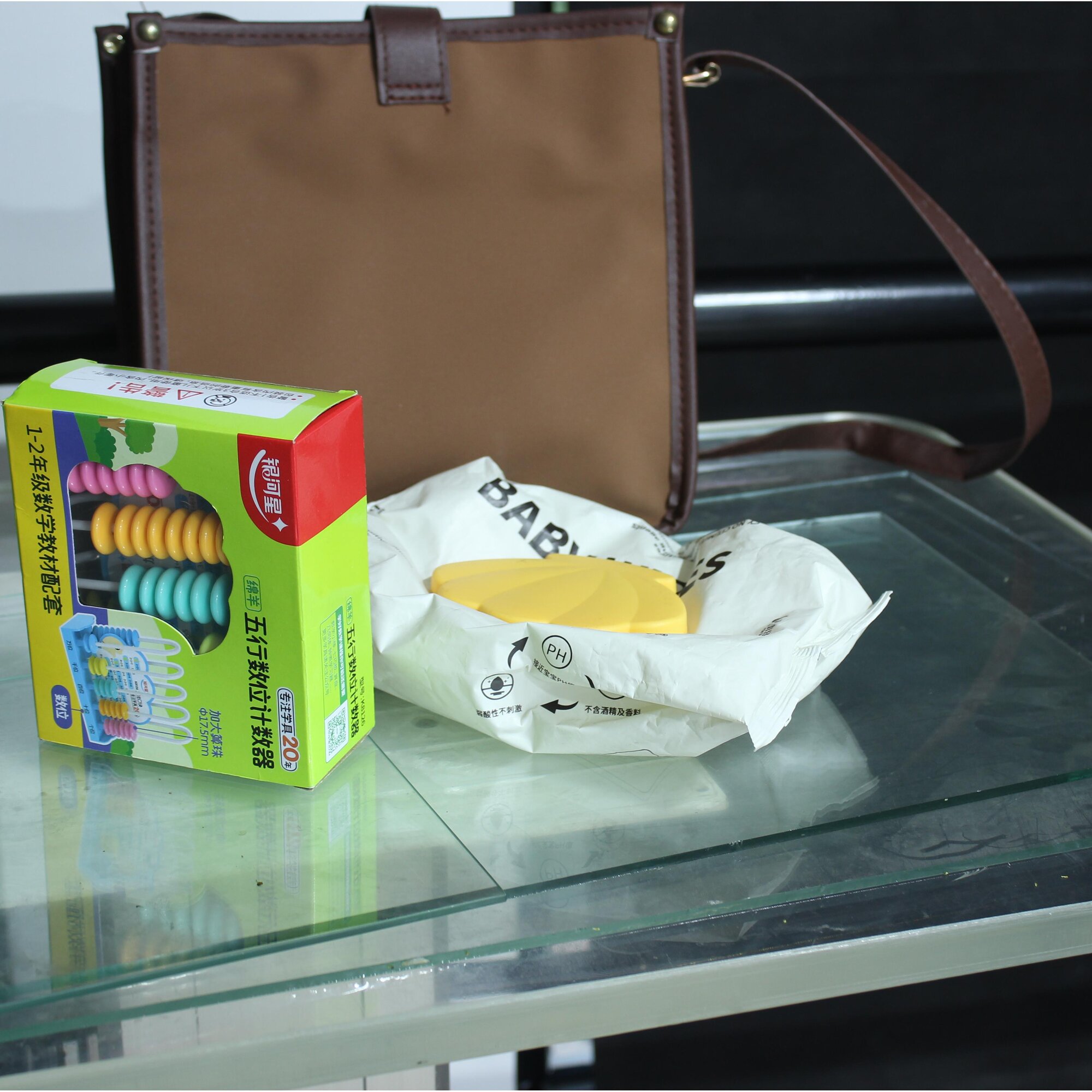} &
            \includegraphics[width=0.23\linewidth]{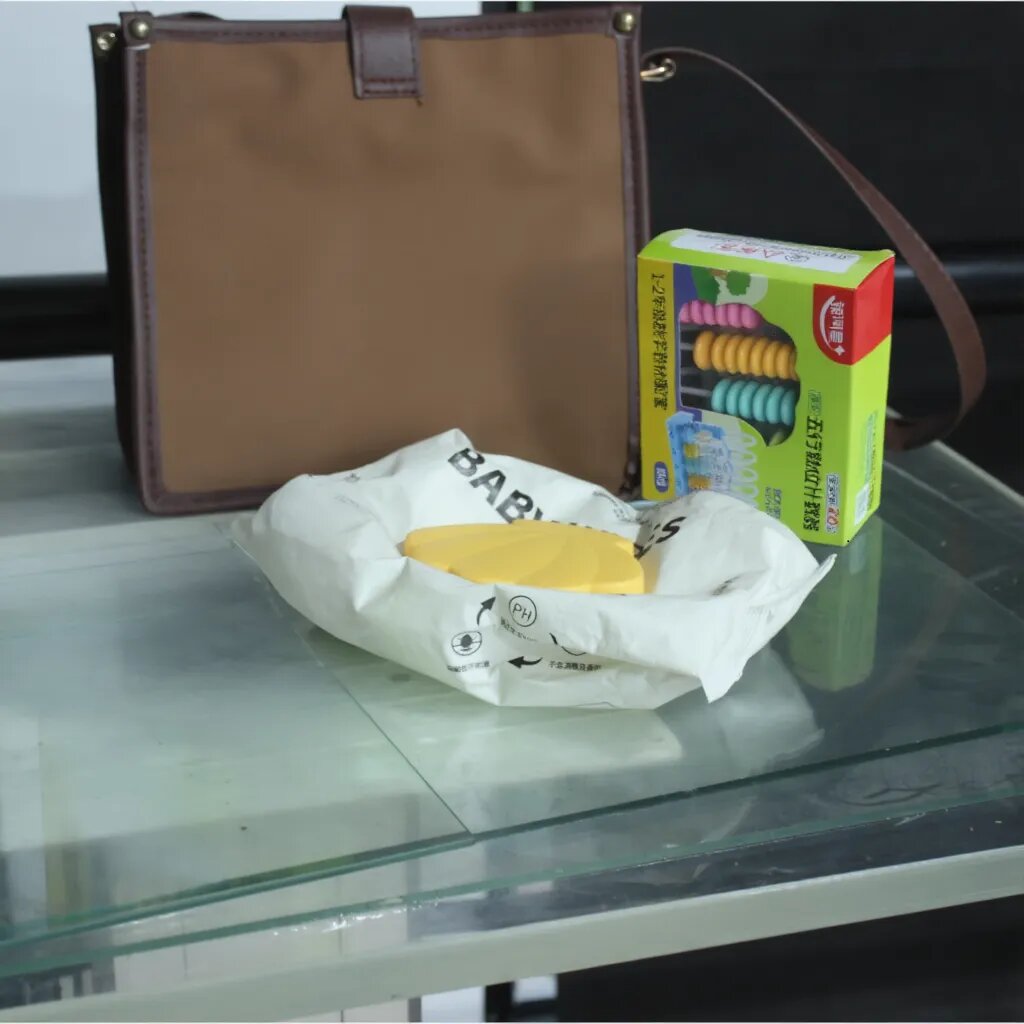} &
            \includegraphics[width=0.23\linewidth]{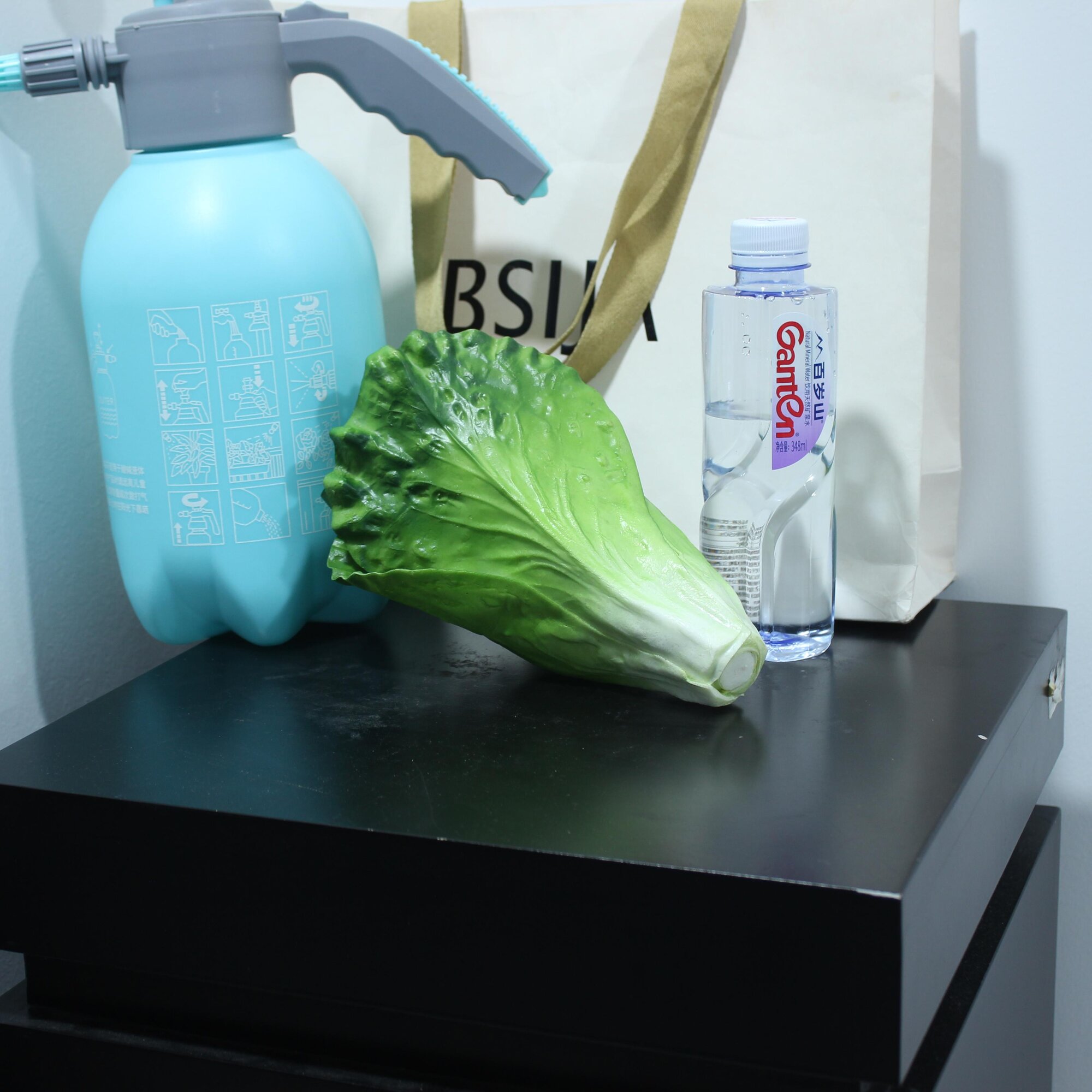} &
            \includegraphics[width=0.23\linewidth]{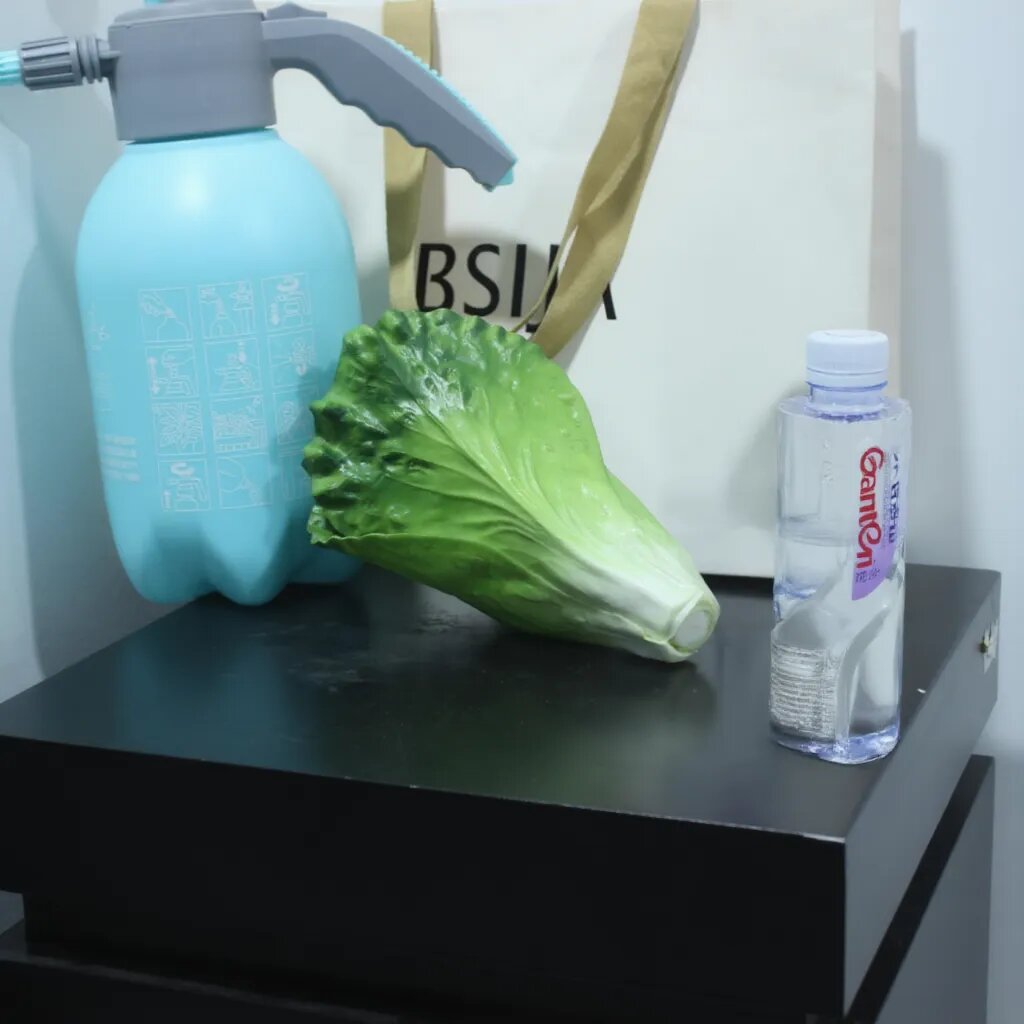}
        \end{tabular}
        \caption{Occlusion}
        \label{fig:occlusion}
    \end{subfigure}

    \begin{subfigure}[b]{\linewidth}
        \centering
        \begin{tabular}{@{}cccc@{}}
            \textbf{Source} & \textbf{Edit} & \textbf{Source} & \textbf{Edit} \\
            \includegraphics[width=0.23\linewidth]{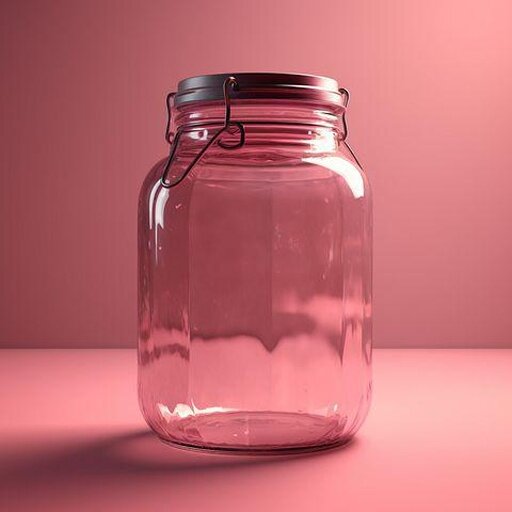} &
            \includegraphics[width=0.23\linewidth]{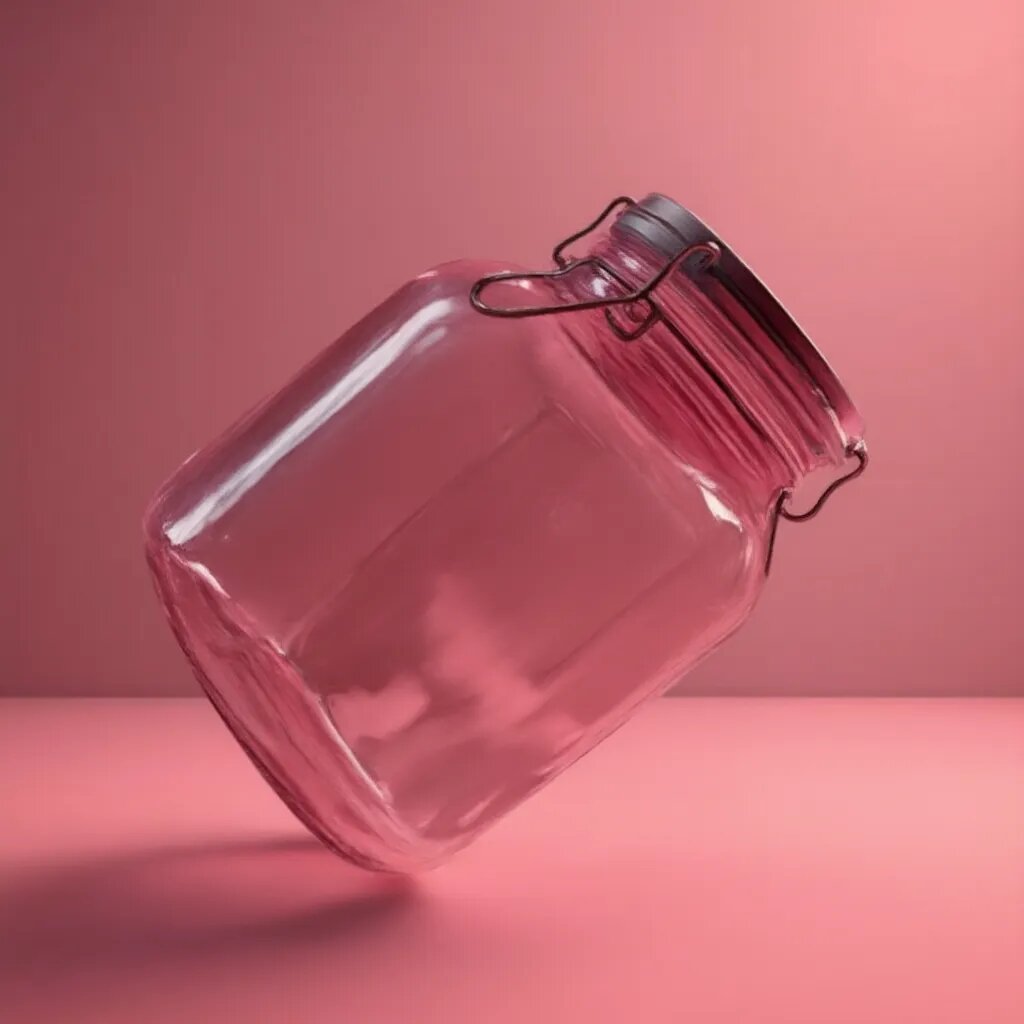} &
            \includegraphics[width=0.23\linewidth]{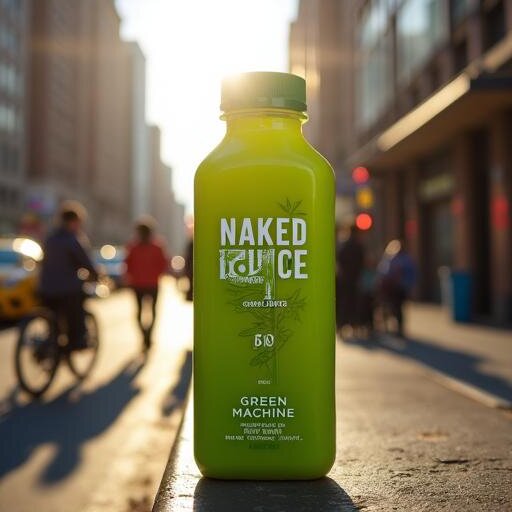} &
            \includegraphics[width=0.23\linewidth]{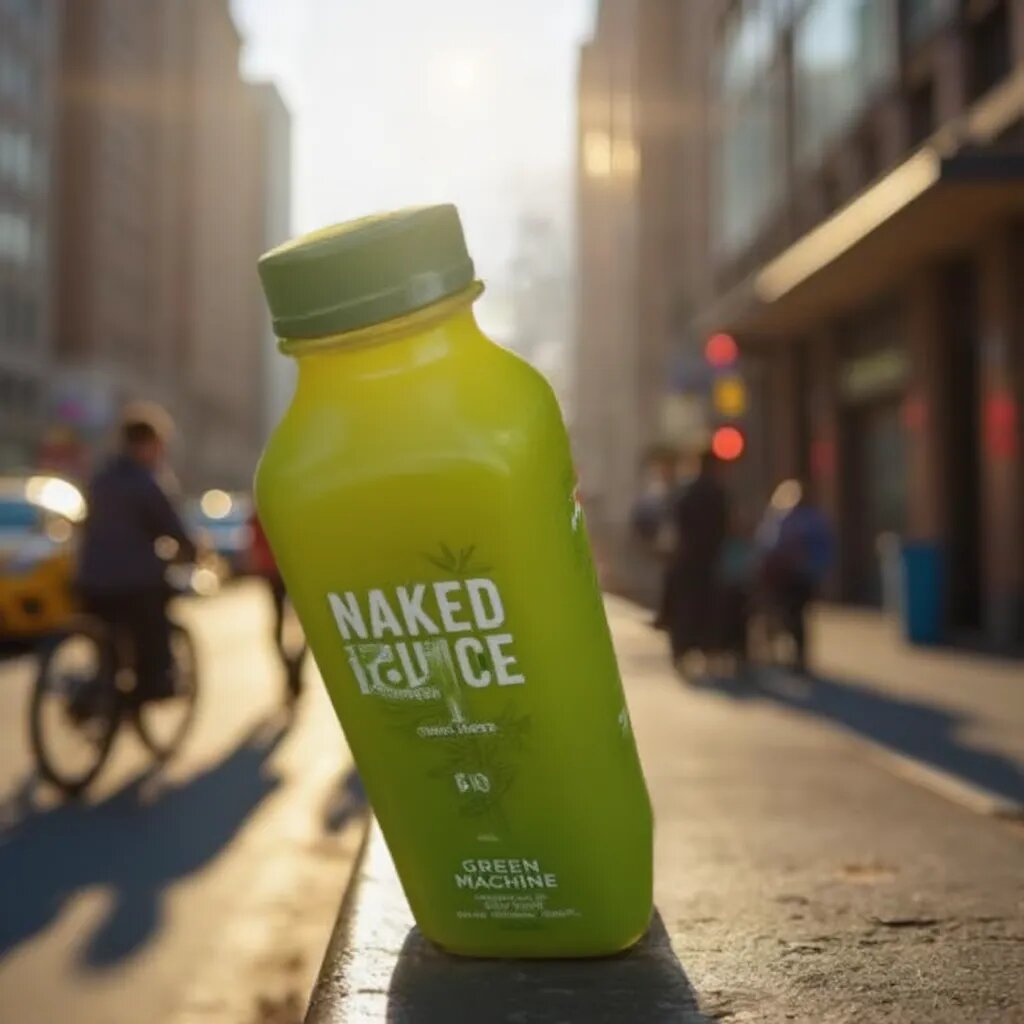}
        \end{tabular}
        \caption{Pitch and roll}
        \label{fig:pitch_roll}
    \end{subfigure}

    \begin{subfigure}[b]{\linewidth}
        \centering
        \begin{tabular}{@{}ccc@{}}
            \textbf{Source} & \textbf{Turn 1} & \textbf{Turn 2} \\
            \includegraphics[width=0.31\linewidth]{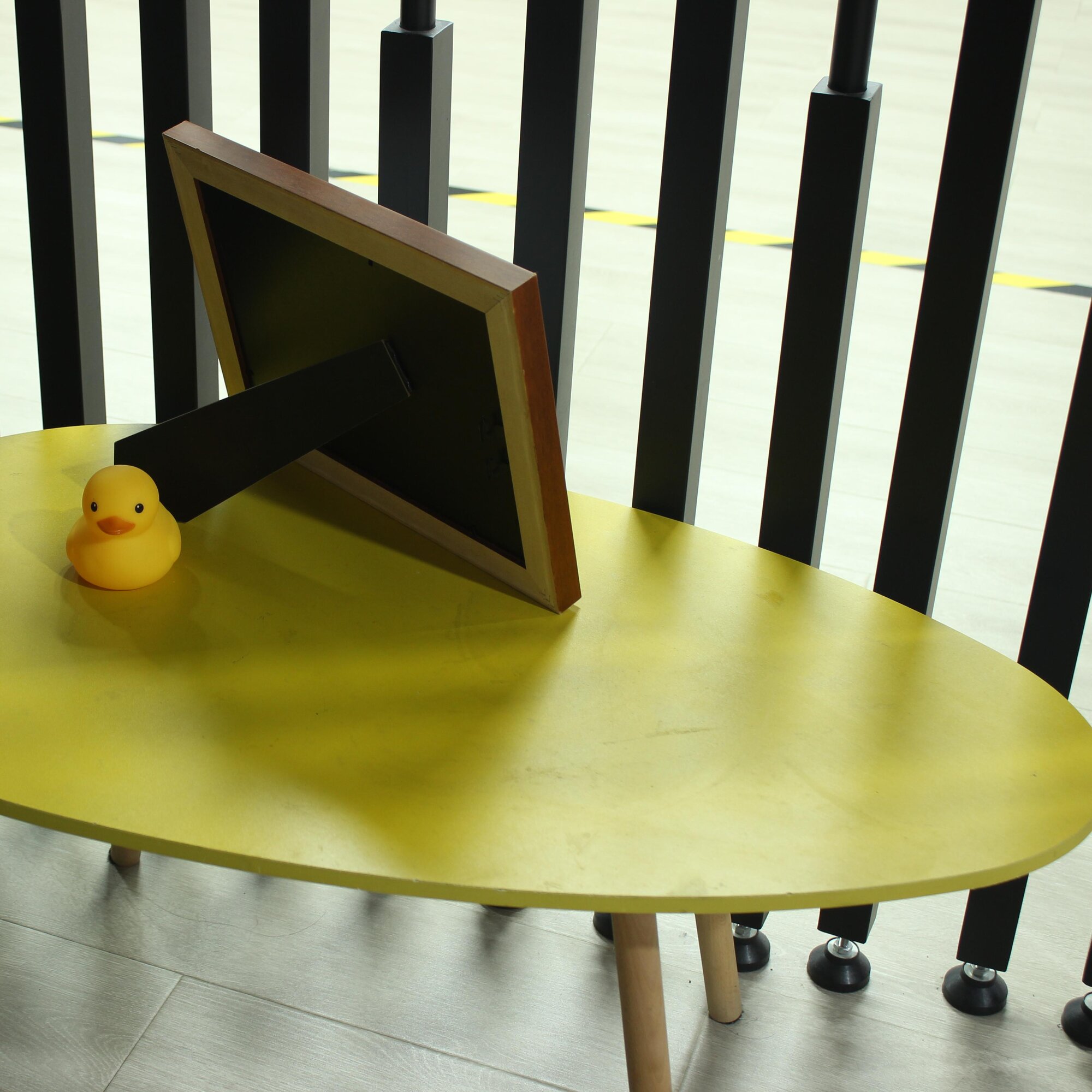} &
            \includegraphics[width=0.31\linewidth]{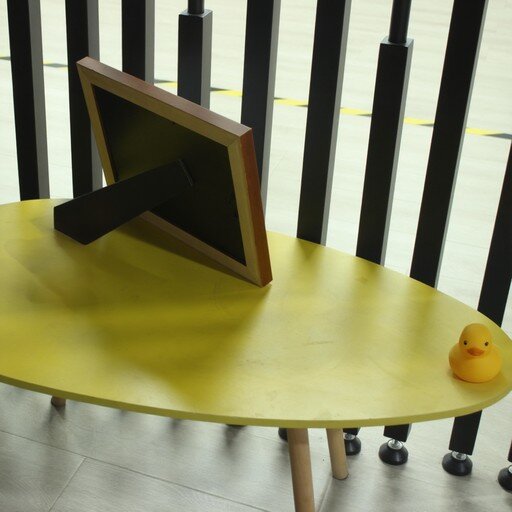} &
            \includegraphics[width=0.31\linewidth]{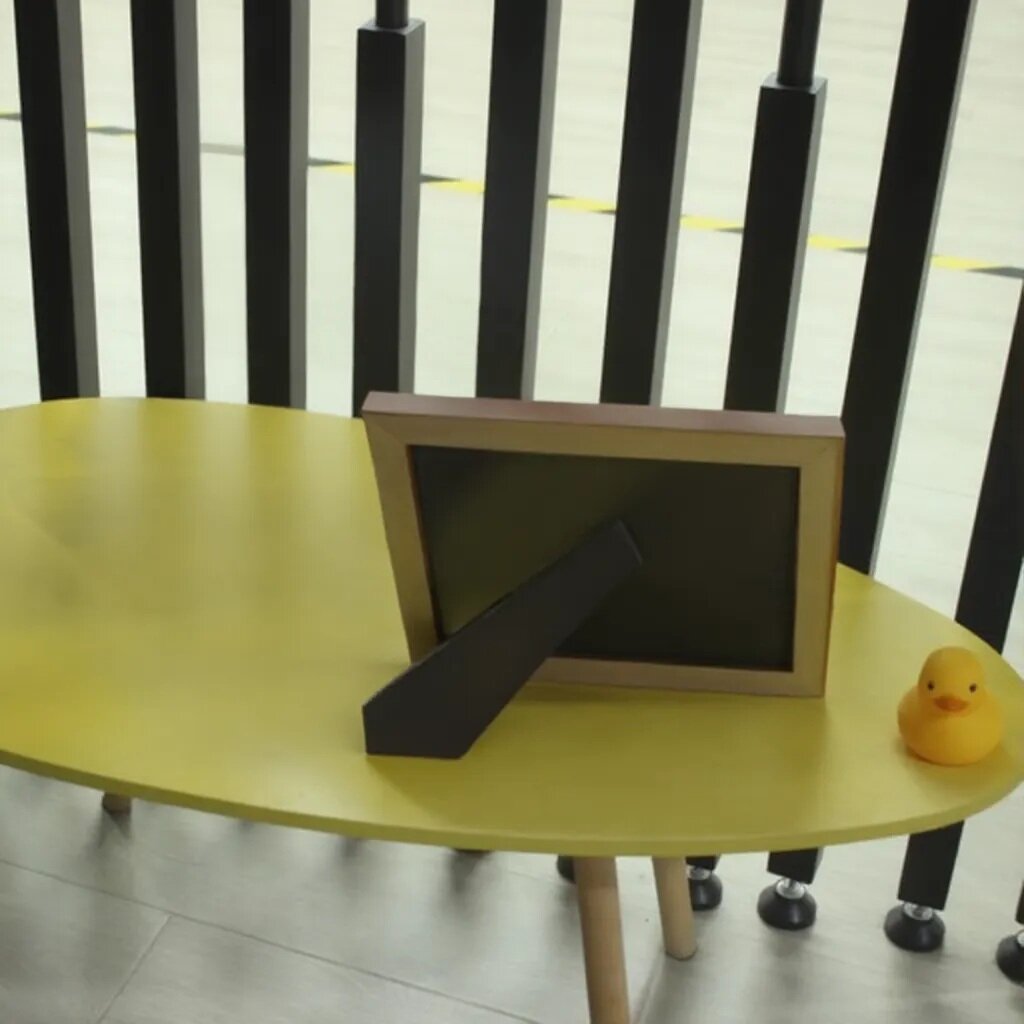}
        \end{tabular}
        \caption{Multi-object}
        \label{fig:multi_object}
    \end{subfigure}

    \caption{\textbf{Qualitative results in challenging scenarios.} Our method demonstrates robustness in complex scenes, enabling precise geometric editing with high visual fidelity. Source images: (a)–(e) and (g) are from ObjectMover Benchmark \cite{yu2025objectmover}; the $1^{\text{st}}$ column of (f) is from PIE-bench \cite{ju2023direct}; and the $3^{\text{rd}}$ column of (f) is from Subject200K \cite{tan2025ominicontrol}.}
    \label{fig:challenging}
\end{figure}

As shown in Fig.~\ref{fig:challenging}, the proposed method demonstrates robustness across several challenging scenarios. It successfully maintains visual coherence under (a) large geometric transformations, (b) reflective surfaces, (c) complex illumination, and (d) accurately editing a specified object in crowded scene structures. Furthermore, the model naturally manages (e) occlusions by preserving physically plausible depth ordering and synthesizing hidden regions. Notably, it achieves zero-shot generalization to unseen (f) pitch and roll rotations, and seamlessly facilitates (g) multi-object editing via an iterative multi-round process.

\section{Conclusion}
In this paper, we introduced a novel 3D-aware visual prompting mechanism utilizing RGB 3D bounding boxes. This design unifies geometric transformations into a single framework, effectively disentangling spatial control from visual appearance. Building upon this mechanism, we devised a two-stage training paradigm. We first address data scarcity via a scalable synthetic pipeline for SFT, establishing fundamental editing capabilities. Subsequently, we bridge the synthetic-to-real domain gap through online RL on unpaired real-world data, guided by a comprehensive joint reward function. Extensive experiments demonstrate that the proposed framework delivers superior performance, successfully balancing precise geometric control with high-fidelity visual preservation.

\begin{acks}
This work was substantially supported by the Innovation and Technology Fund (ITF) of the Innovation and Technology Commission (ITC) of the Hong Kong Special Administrative Region (HKSAR) Government [Project No. ITS/269/24FP].
\end{acks}

\bibliographystyle{ACM-Reference-Format}
\bibliography{ref}

\clearpage

\begin{figure}[htbp]
    \centering
    \setlength{\tabcolsep}{2pt}        
    \renewcommand{\arraystretch}{0.9}  

    \begin{tabular}{@{}ccccccc@{}}
        {\small \textbf{Src}} &
        {\small \textbf{Src Box}} &
        {\small \textbf{GT}} &
        {\small \textbf{GT Box}} &
        {\small \textbf{(A)}} &
        {\small \textbf{(B)}} &
        {\small \textbf{(C)}} 
        \\[3pt]

        \includegraphics[width=0.12\linewidth]{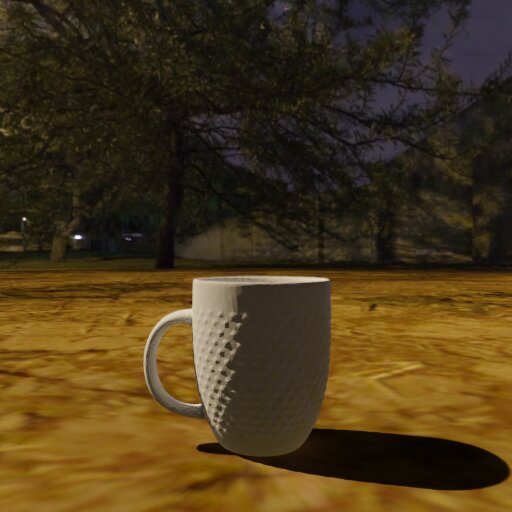} &

        \includegraphics[width=0.12\linewidth]{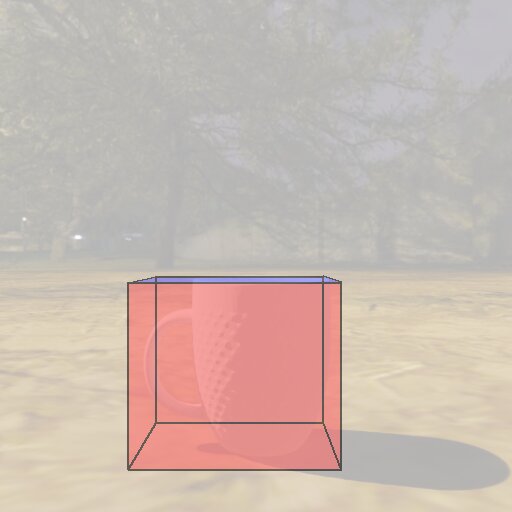} &

        \includegraphics[width=0.12\linewidth]{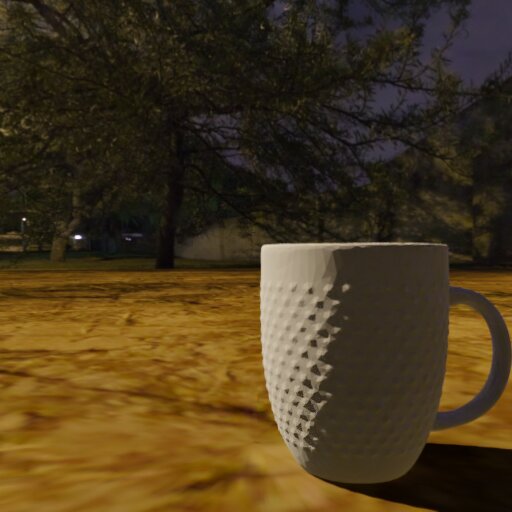} &

        \includegraphics[width=0.12\linewidth]{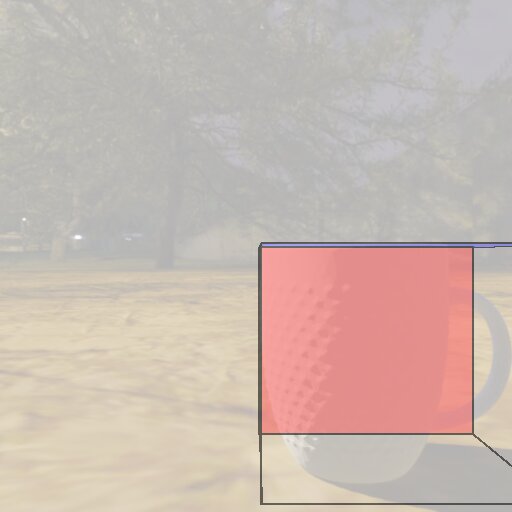} &

        \includegraphics[width=0.12\linewidth]{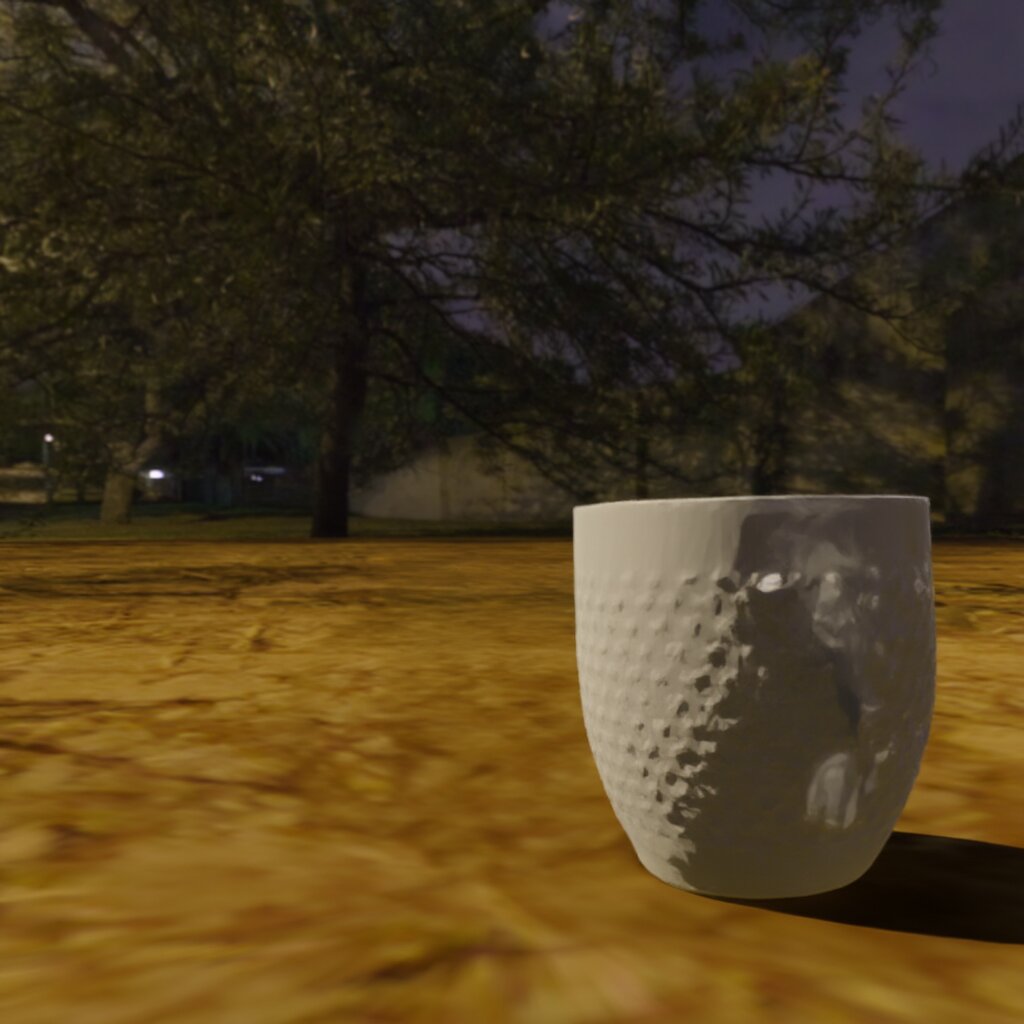} &

        \includegraphics[width=0.12\linewidth]{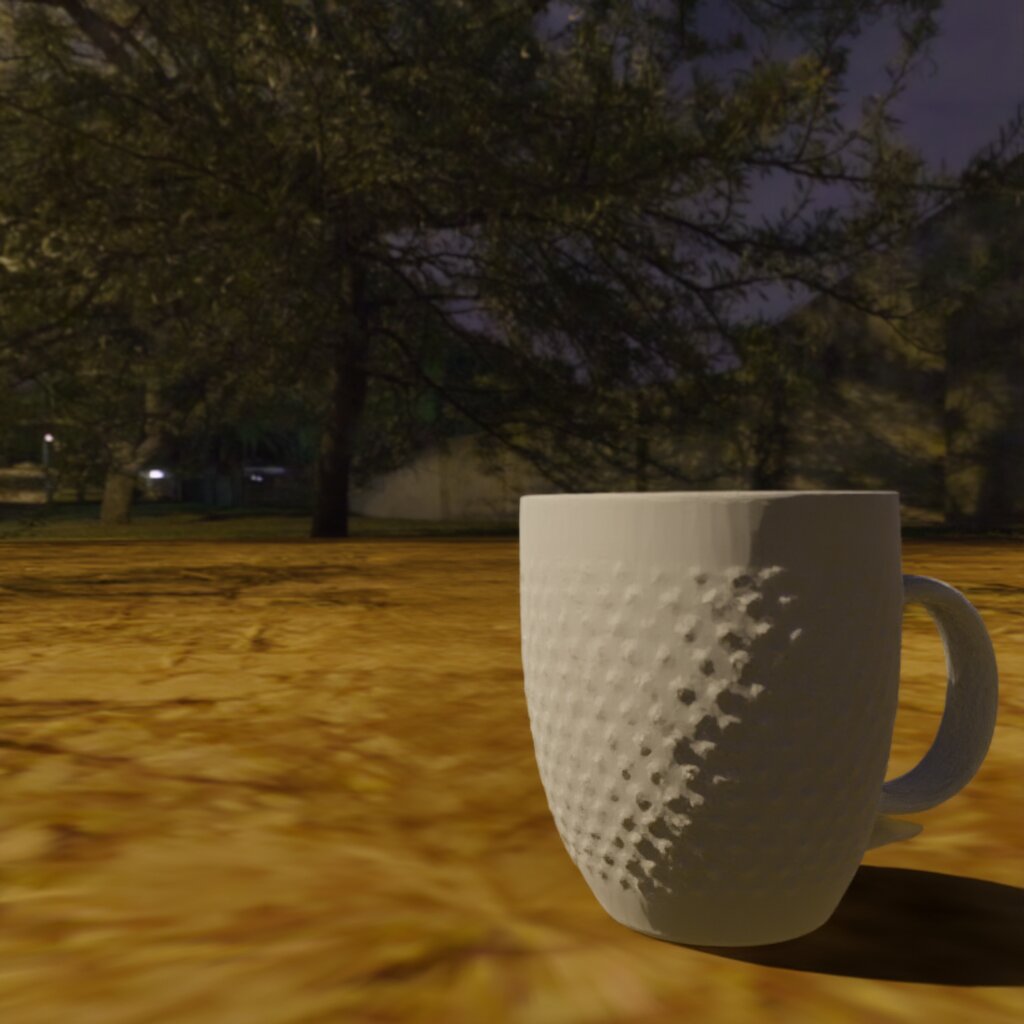} &
        
        \includegraphics[width=0.12\linewidth]{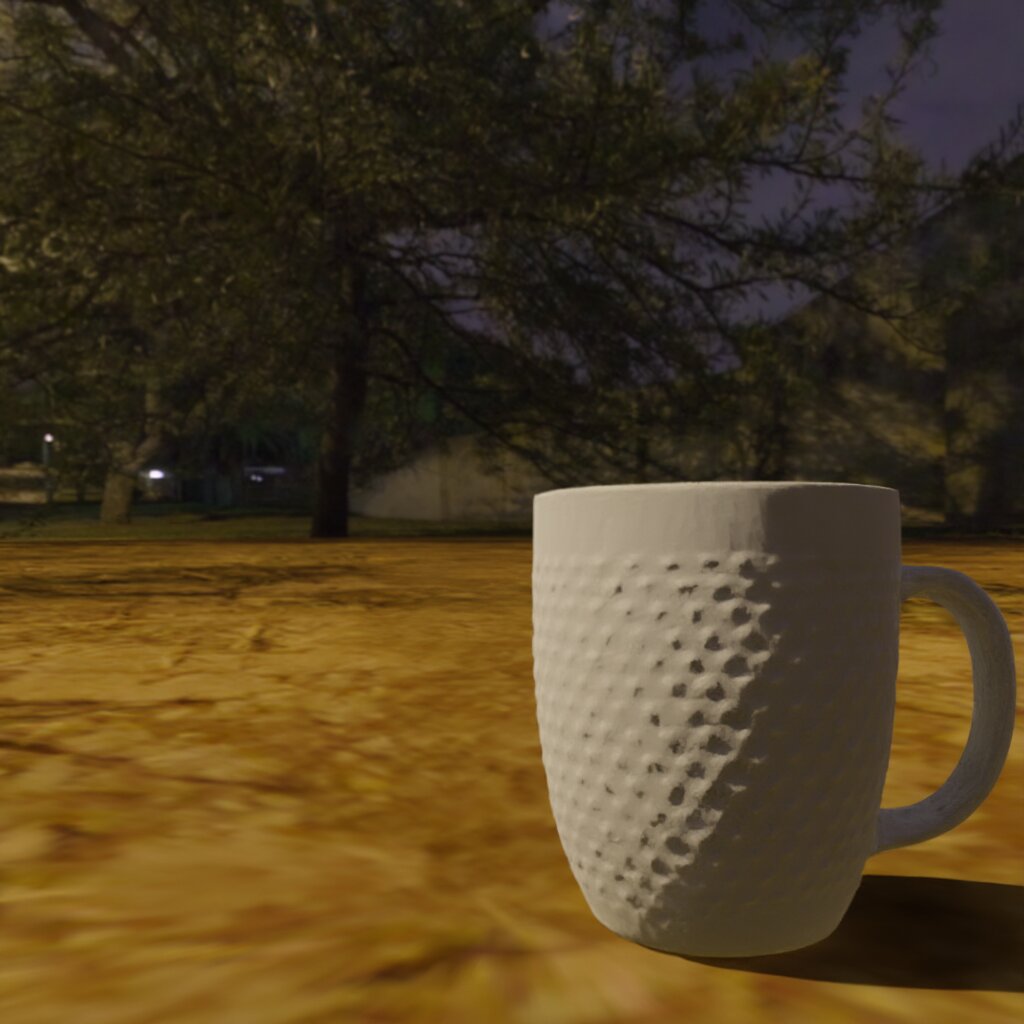}
    \end{tabular}

    \caption{\textbf{Qualitative comparison under different conditions.} Left to right: source image, source RGB 3D bbox, ground truth, and ground truth RGB 3D bbox. Results are shown for: (A) text prompting only, (B) RGB 3D bbox only, and (C) RGB 3D bbox + text.}

    \label{fig:condition_type}
\end{figure}

\begin{figure}[htbp]
    \centering

    \setlength{\tabcolsep}{3pt} 
    \begin{tabular}{@{}cccc@{}}
        \textbf{Src} & \textbf{(a)} & \textbf{(b)} & \textbf{(c)} \\[2pt] 
        \includegraphics[width=0.1\textwidth]{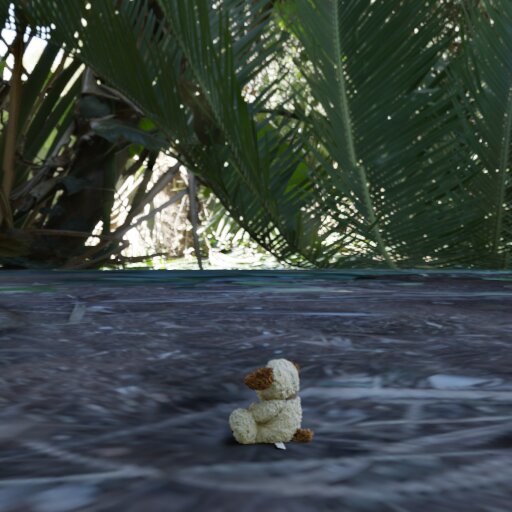} &
        \includegraphics[width=0.1\textwidth]{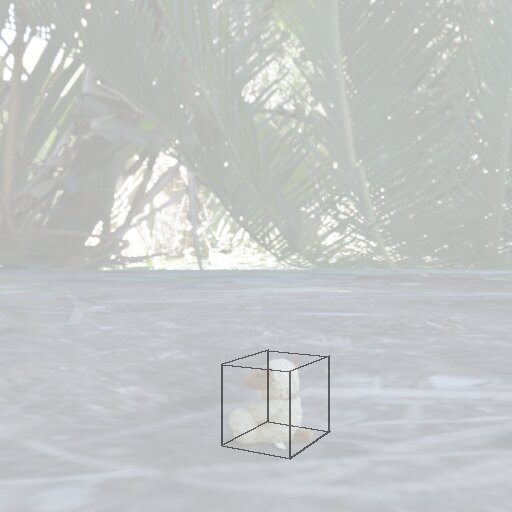} &
        \includegraphics[width=0.1\textwidth]{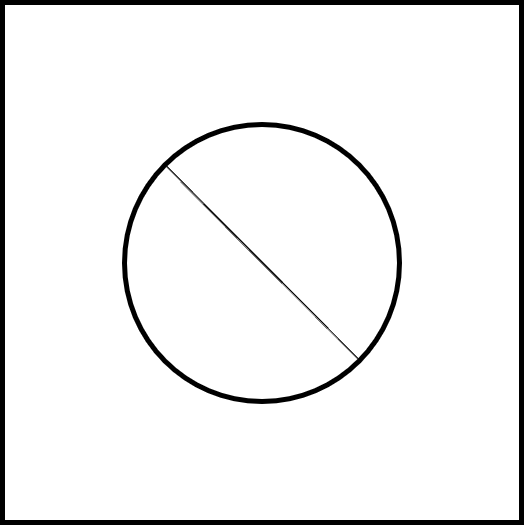} &
        \includegraphics[width=0.1\textwidth]{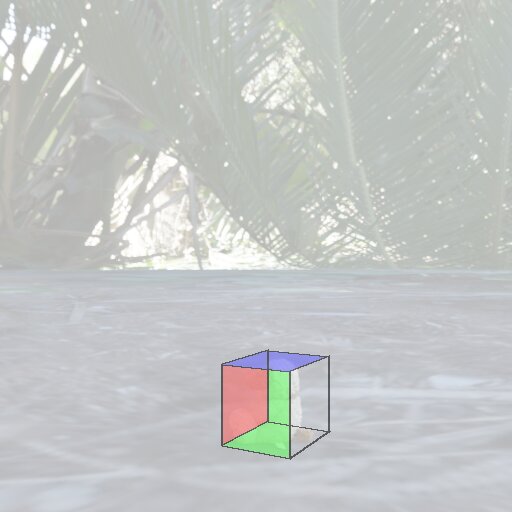} \\[1pt]

        &
        \includegraphics[width=0.1\textwidth]{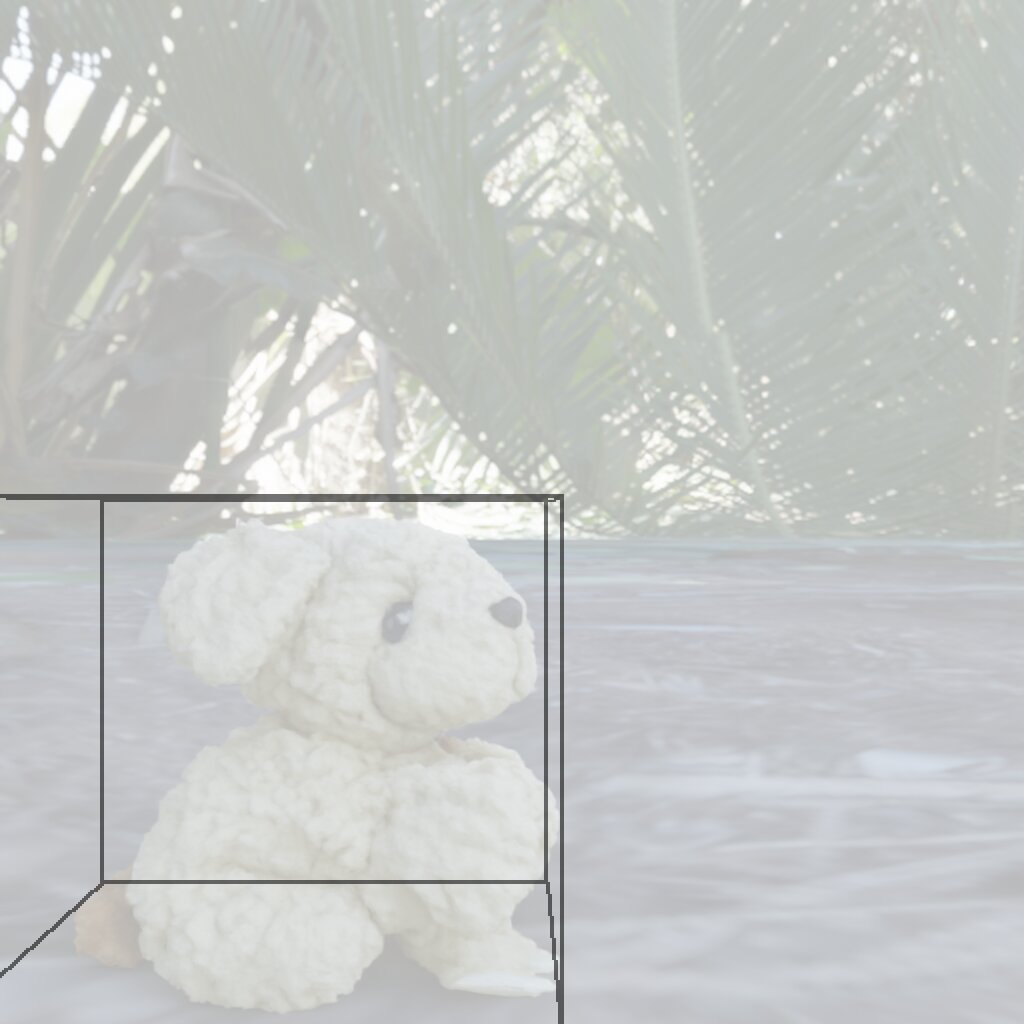} &
        \includegraphics[width=0.1\textwidth]{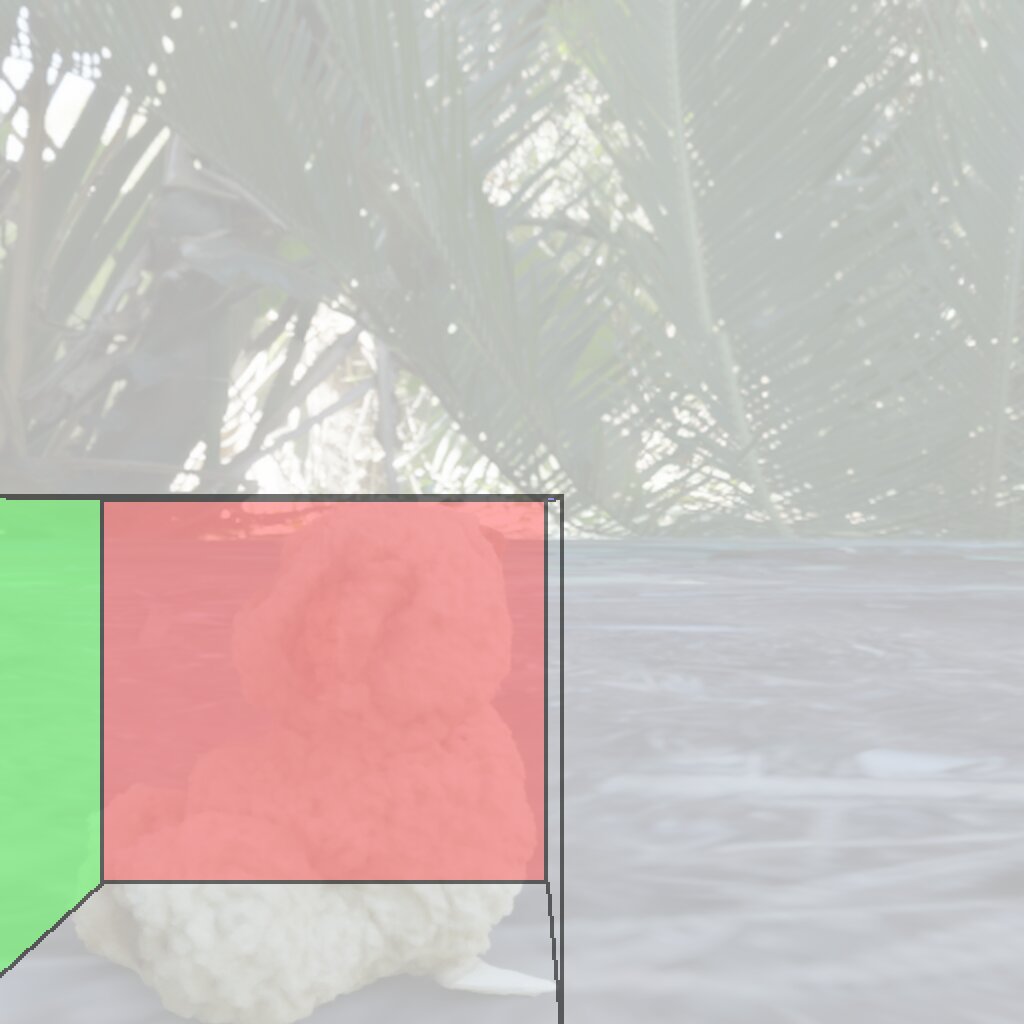} &
        \includegraphics[width=0.1\textwidth]{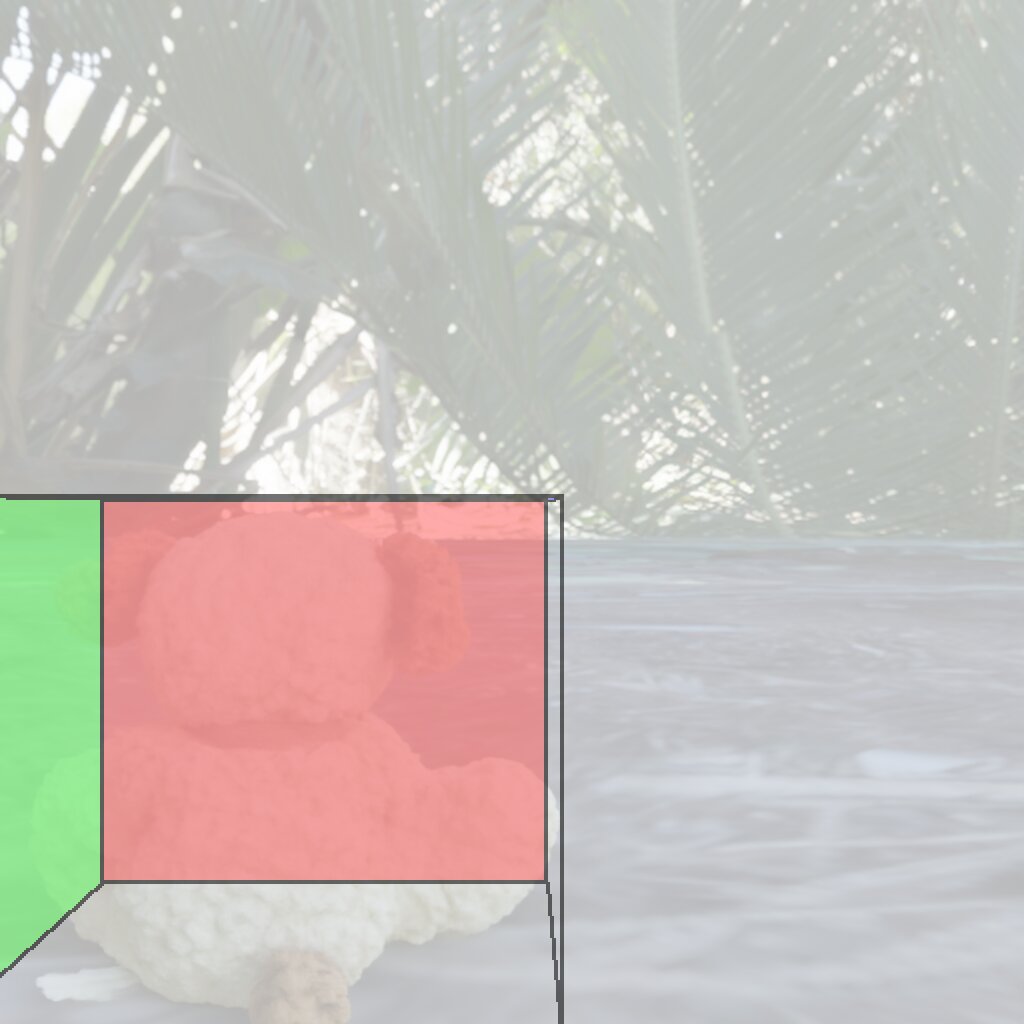} \\[-6pt]
        \textbf{GT} & & & \\[-2pt] 
        
        \includegraphics[width=0.1\textwidth]{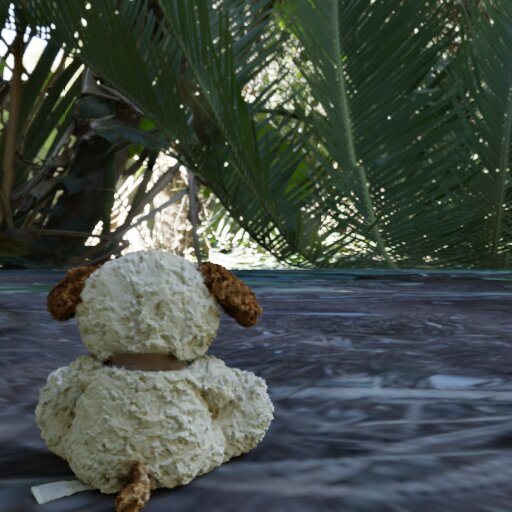} &
        \includegraphics[width=0.1\textwidth]{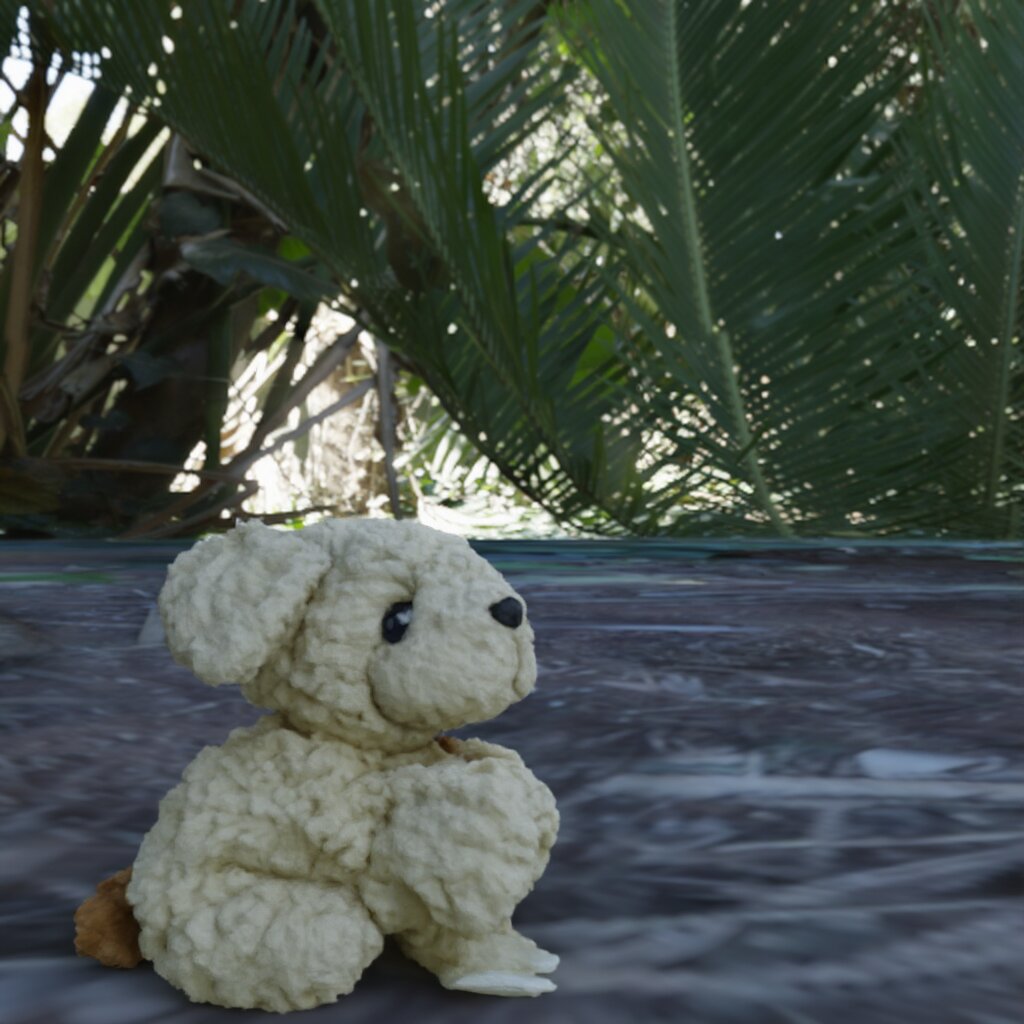} &
        \includegraphics[width=0.1\textwidth]{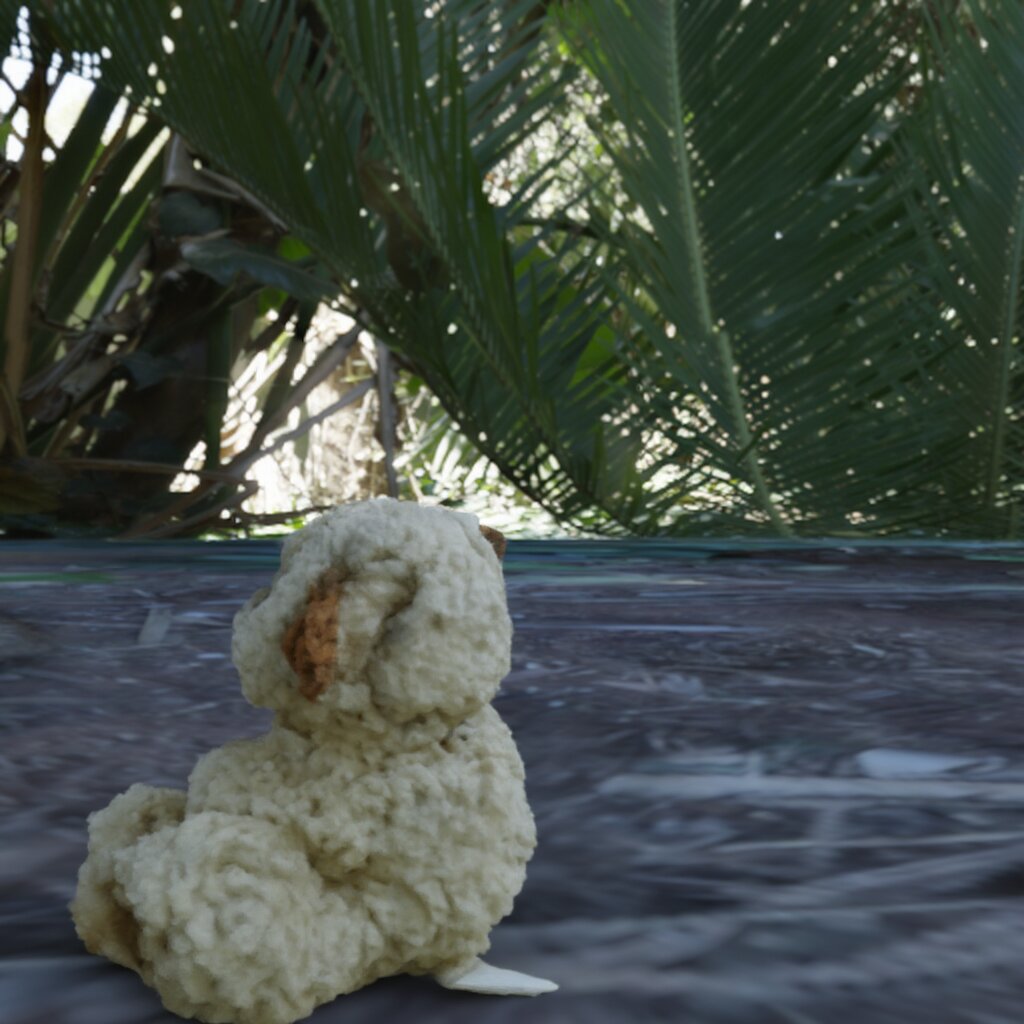} &
        \includegraphics[width=0.1\textwidth]{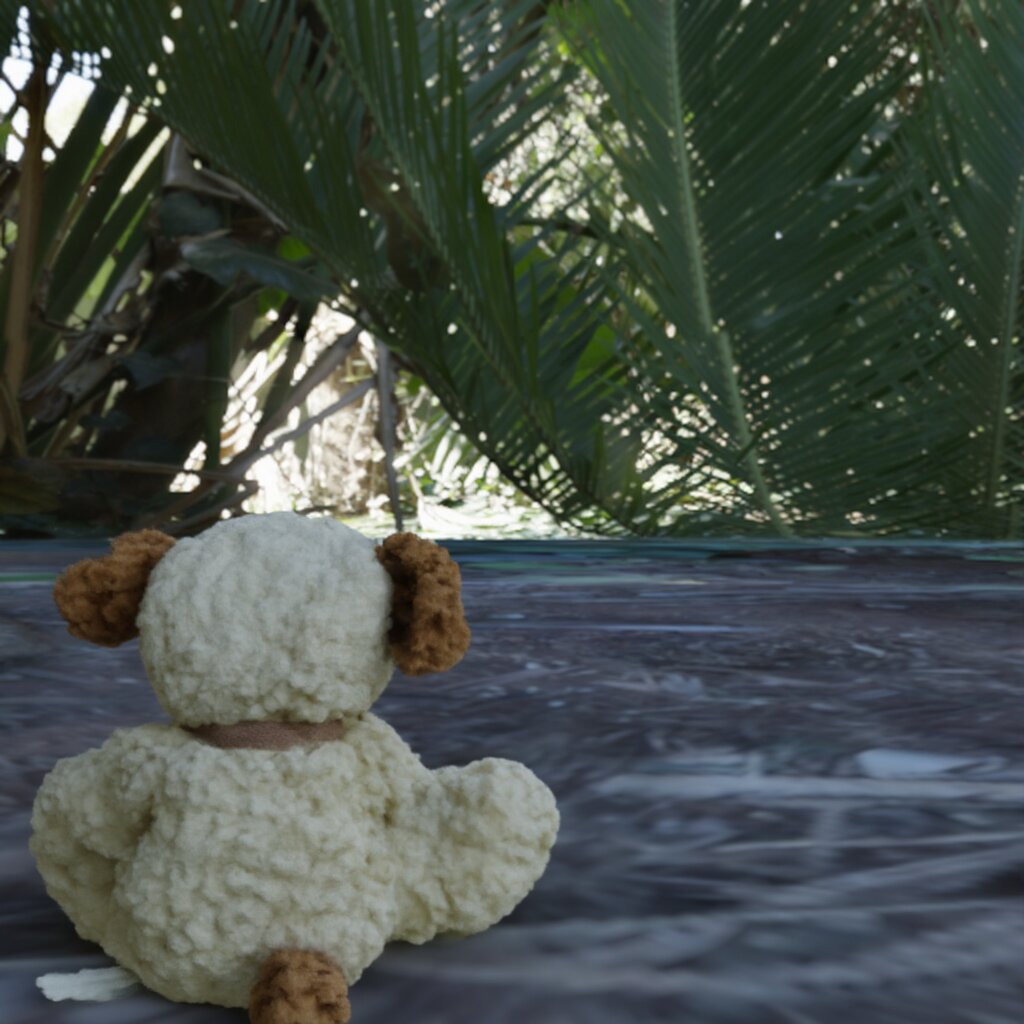} 
    \end{tabular}

    \caption{\textbf{Qualitative comparison under different visual prompting conditions.} Cols (a-c): dual gray, single RGB, and dual RGB bboxes. Top rows: conditions with source (Src). Bottom: results with ground truth (GT).}
    \label{fig:ablation_box}
\end{figure}

\begin{figure}[htbp]
    \centering
    \setlength{\tabcolsep}{2pt}       
    \renewcommand{\arraystretch}{0.9}  

    \begin{tabular}{@{}cccc@{}}
        {\small \textbf{Src}} &
        {\small \textbf{GT}} &
        {\small \textbf{SFT}} &
        {\small \textbf{SFT + RL}} \\[3pt]
        \includegraphics[width=0.24\linewidth]{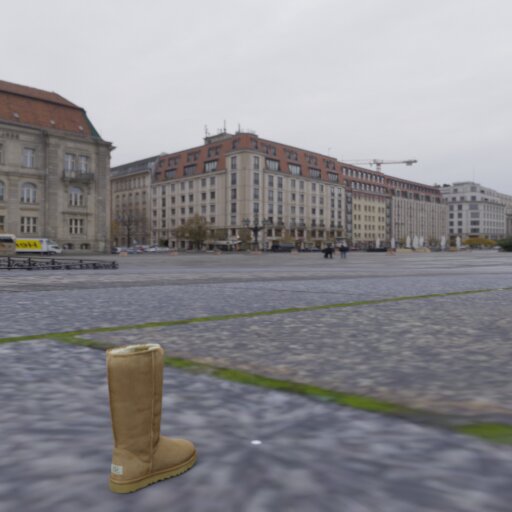} &
        \includegraphics[width=0.24\linewidth]{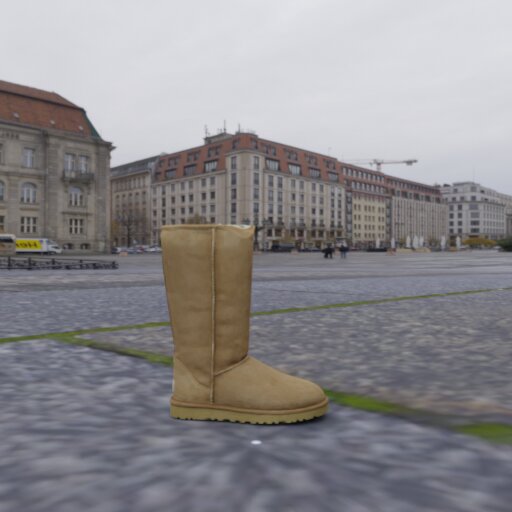} &
        \includegraphics[width=0.24\linewidth]{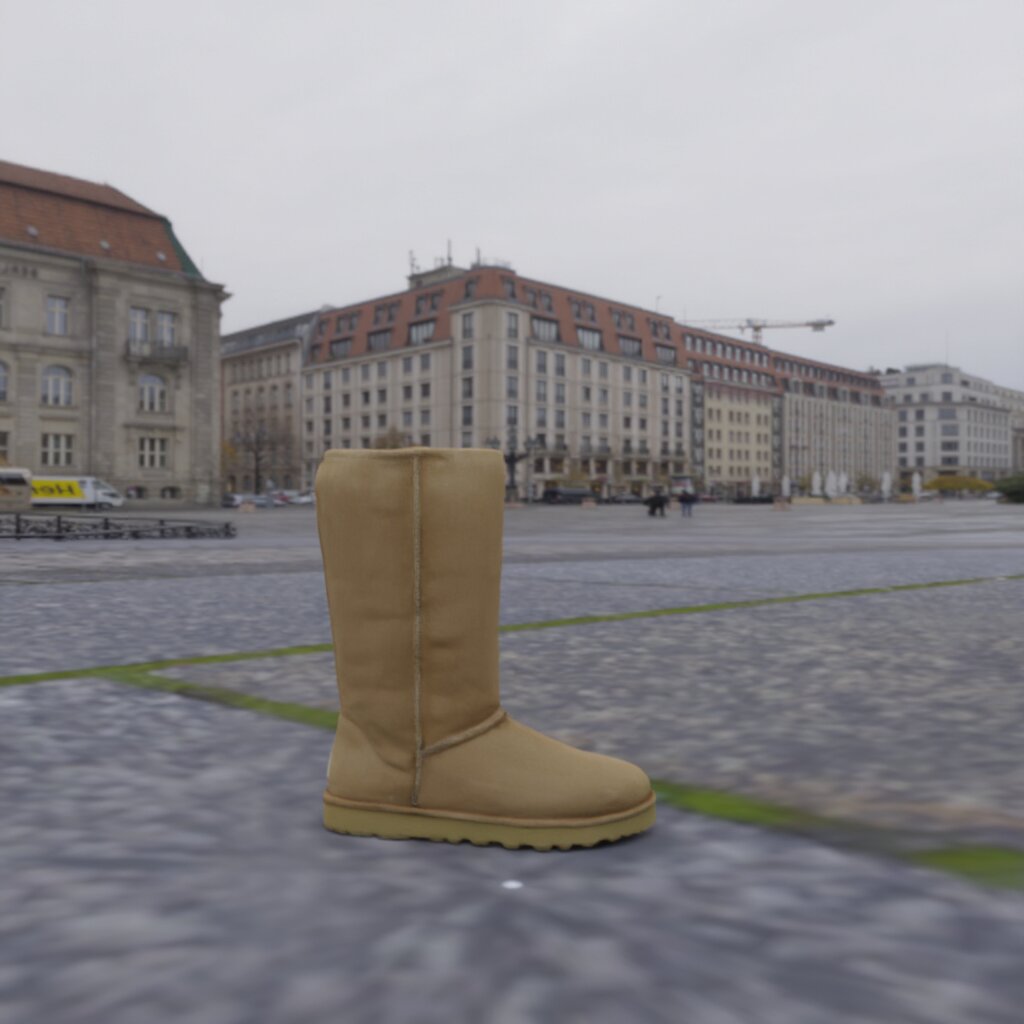} &
        \includegraphics[width=0.24\linewidth]{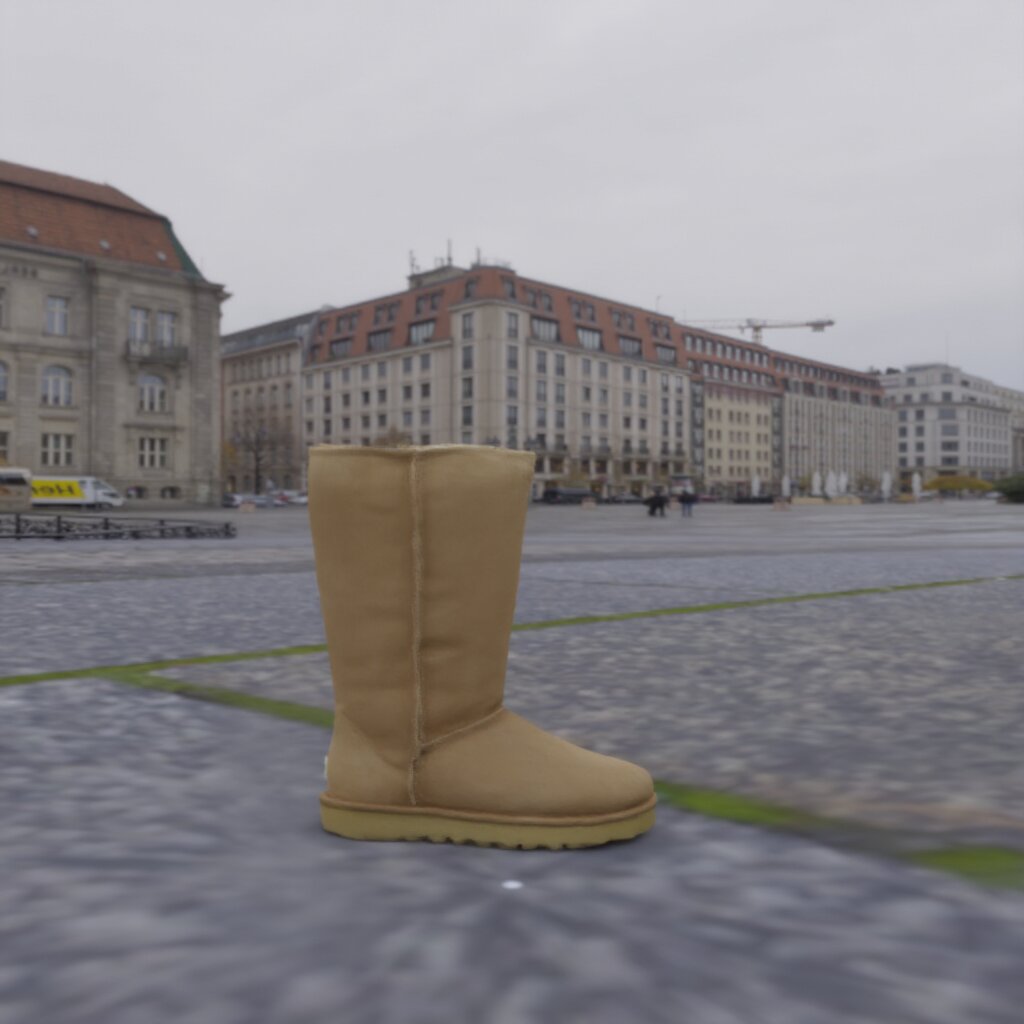}
    \end{tabular}

    \caption{\textbf{Qualitative comparison of SFT and SFT+RL on synthetic data.} Left to right: source image, ground truth, SFT, and SFT+RL.}
    \label{fig:sft_rl_syn}
\end{figure}

\begin{figure}[htbp]
    \centering
    \setlength{\tabcolsep}{1pt}
    \renewcommand{\arraystretch}{0.9}

    \begin{tabular}{@{}cccccc@{}}

        {\scriptsize \textbf{Src}} &
        {\scriptsize \textbf{SFT}} &
        {\scriptsize \textbf{w/o $r_{\text{rot}}$}} &
        {\scriptsize \textbf{w/o $r_{\text{ts}}$}} &
        {\scriptsize \textbf{w/o $r_{\text{sim}}$ }} &
        {\scriptsize \textbf{Ours}} \\[2pt]

        \includegraphics[width=0.16\linewidth]{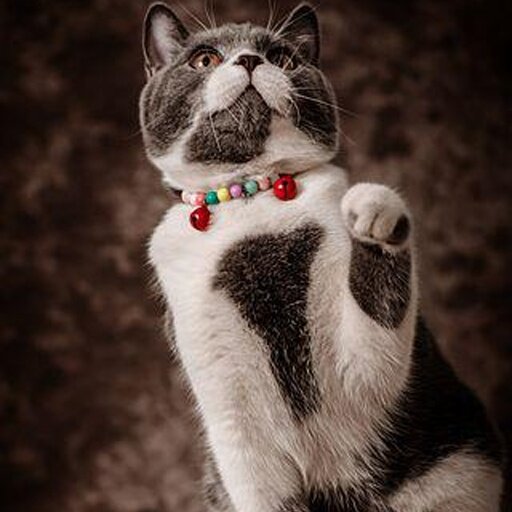} &
    
        \includegraphics[width=0.16\linewidth]{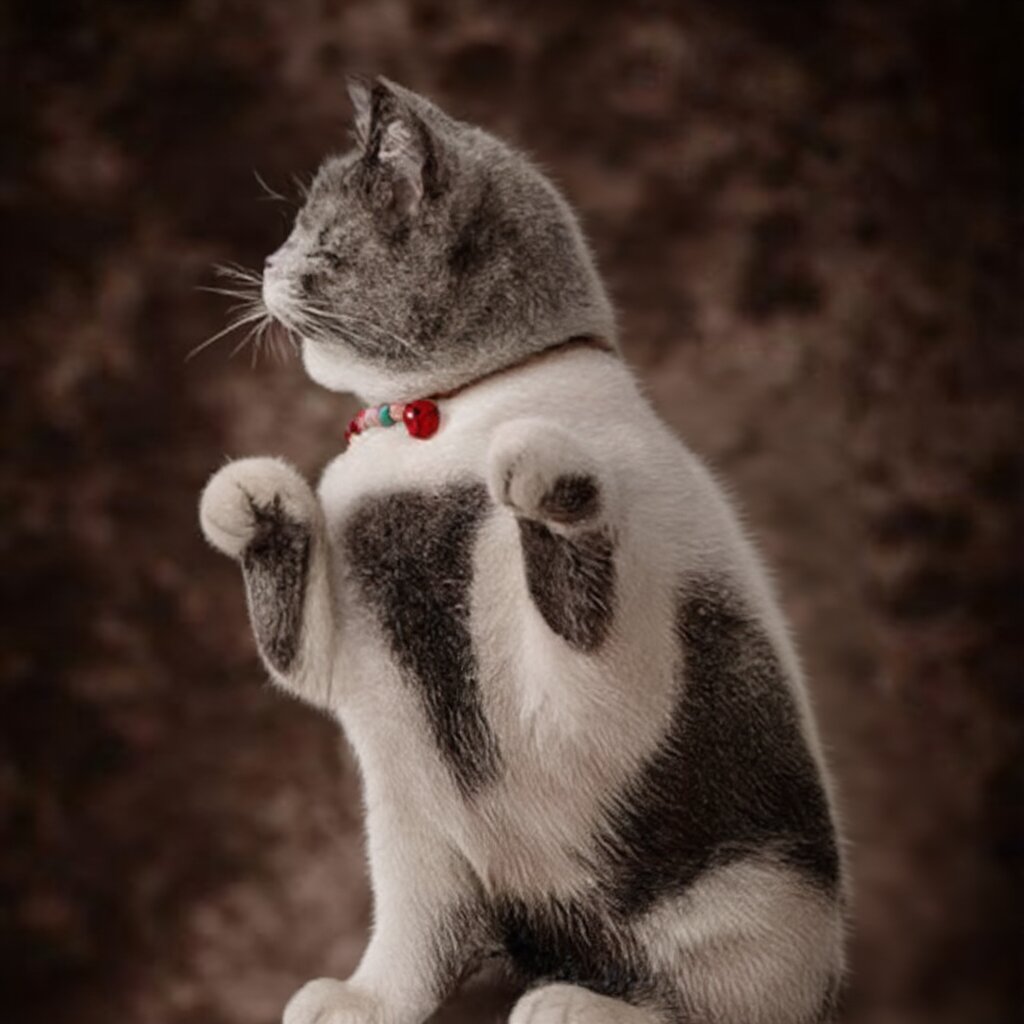} &
        \includegraphics[width=0.16\linewidth]{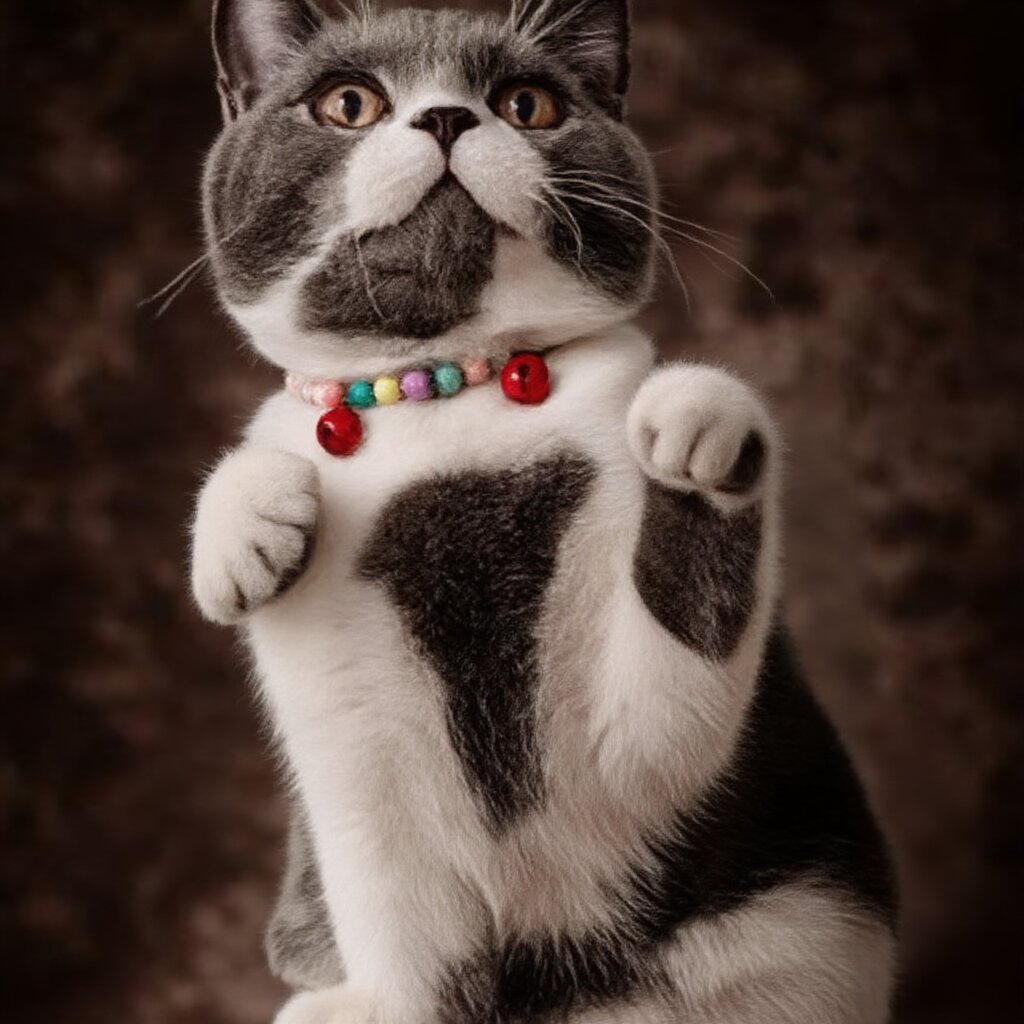} &
        \includegraphics[width=0.16\linewidth]{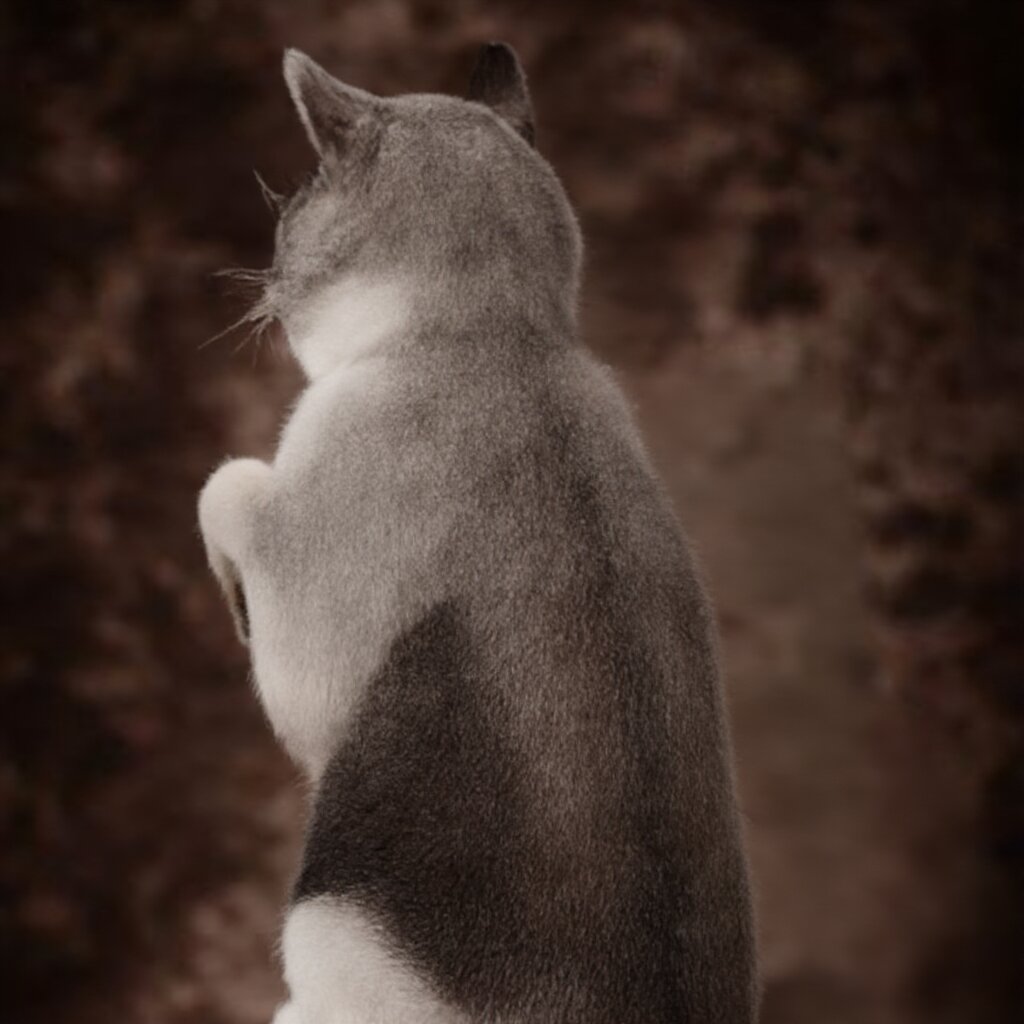} &
        \includegraphics[width=0.16\linewidth]{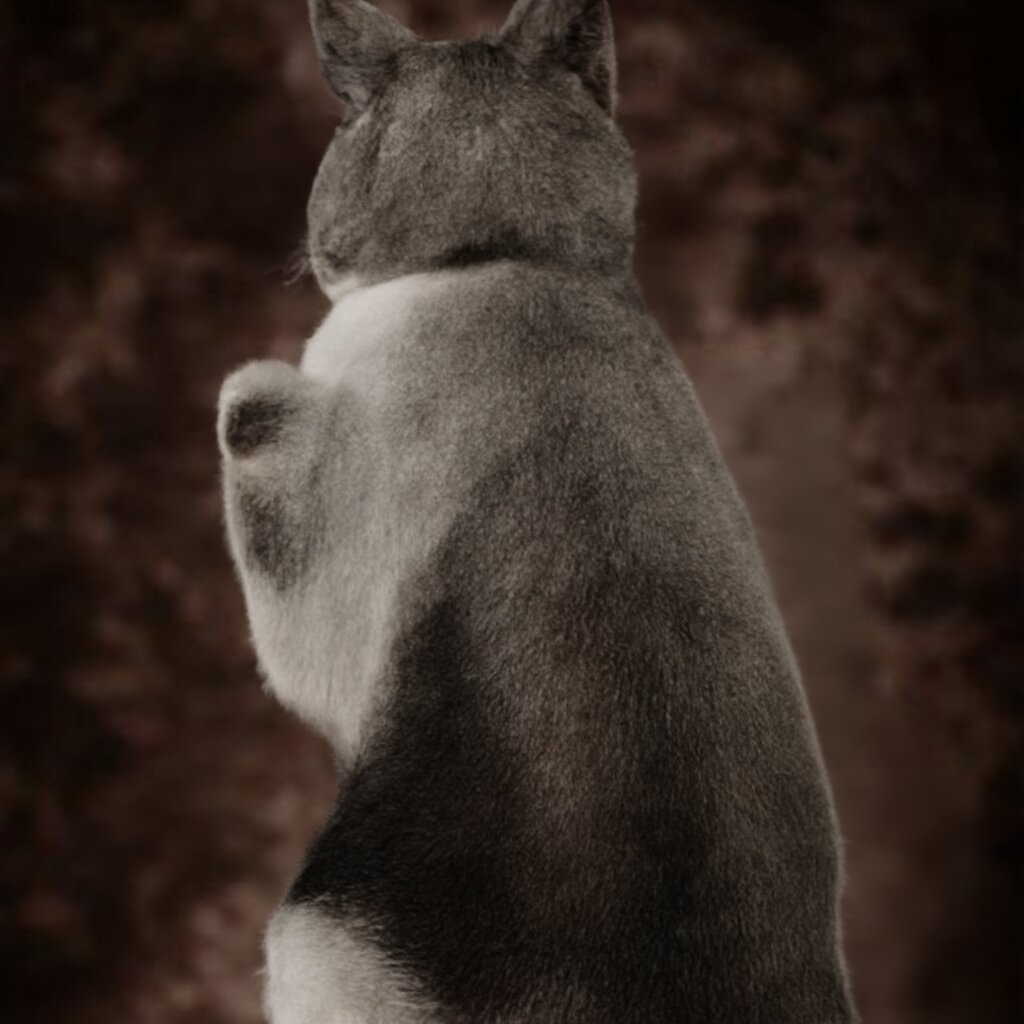} &
        
        \includegraphics[width=0.16\linewidth]{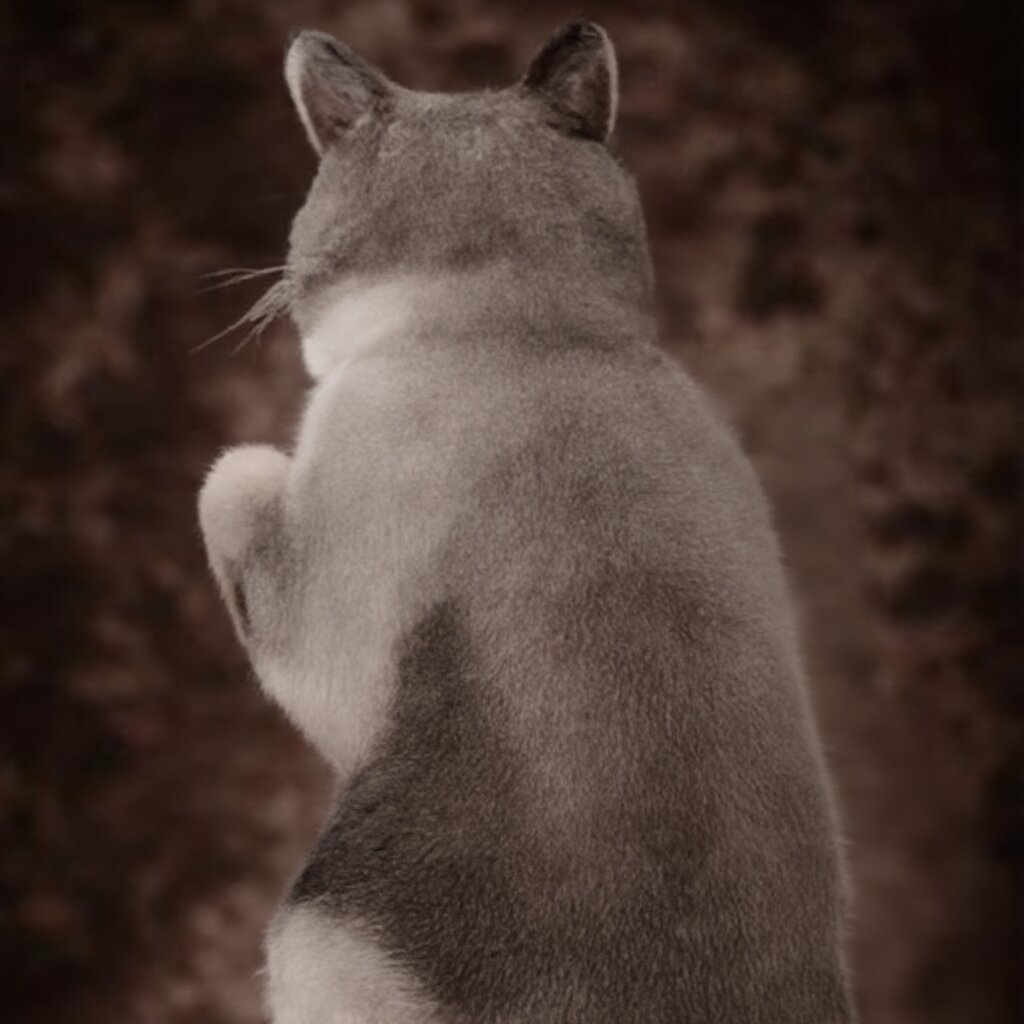} \\[2pt]

        \multicolumn{6}{c}{\scriptsize \textit{``Turn left 150.0 degrees, shrink by a factor of 0.90, move left 0.10."}.} \\
        
        \includegraphics[width=0.16\linewidth]{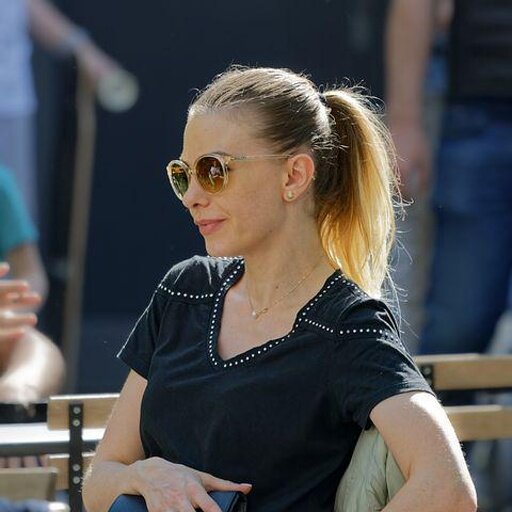} &

        \includegraphics[width=0.16\linewidth]{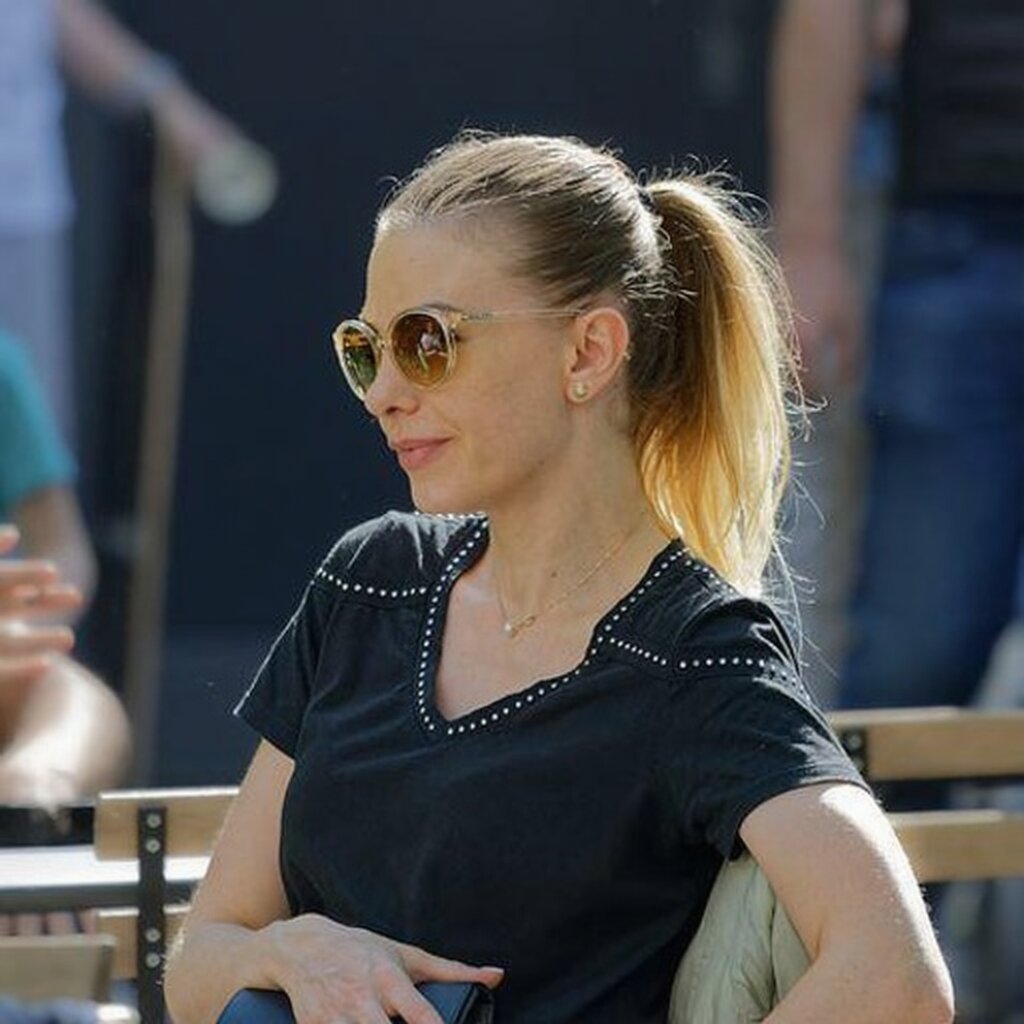} &

        \includegraphics[width=0.16\linewidth]{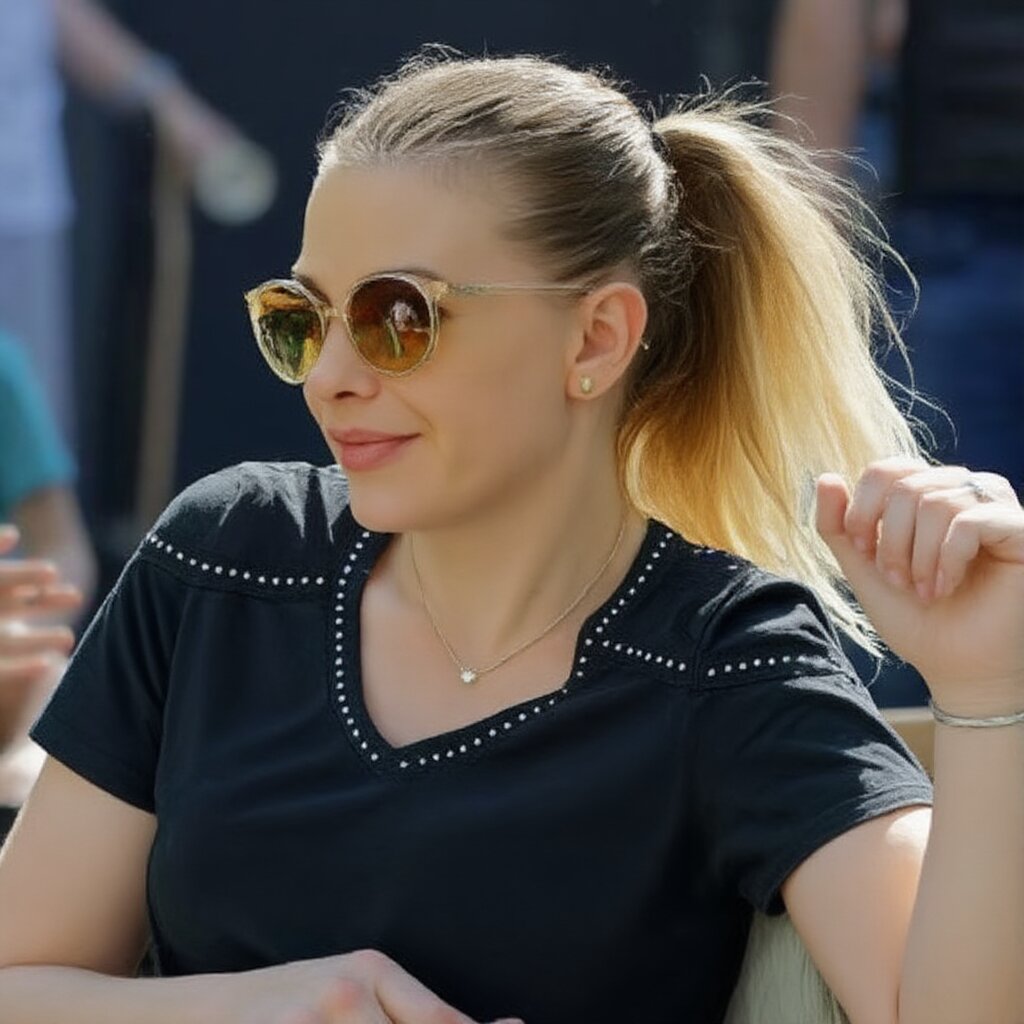} &

        \includegraphics[width=0.16\linewidth]{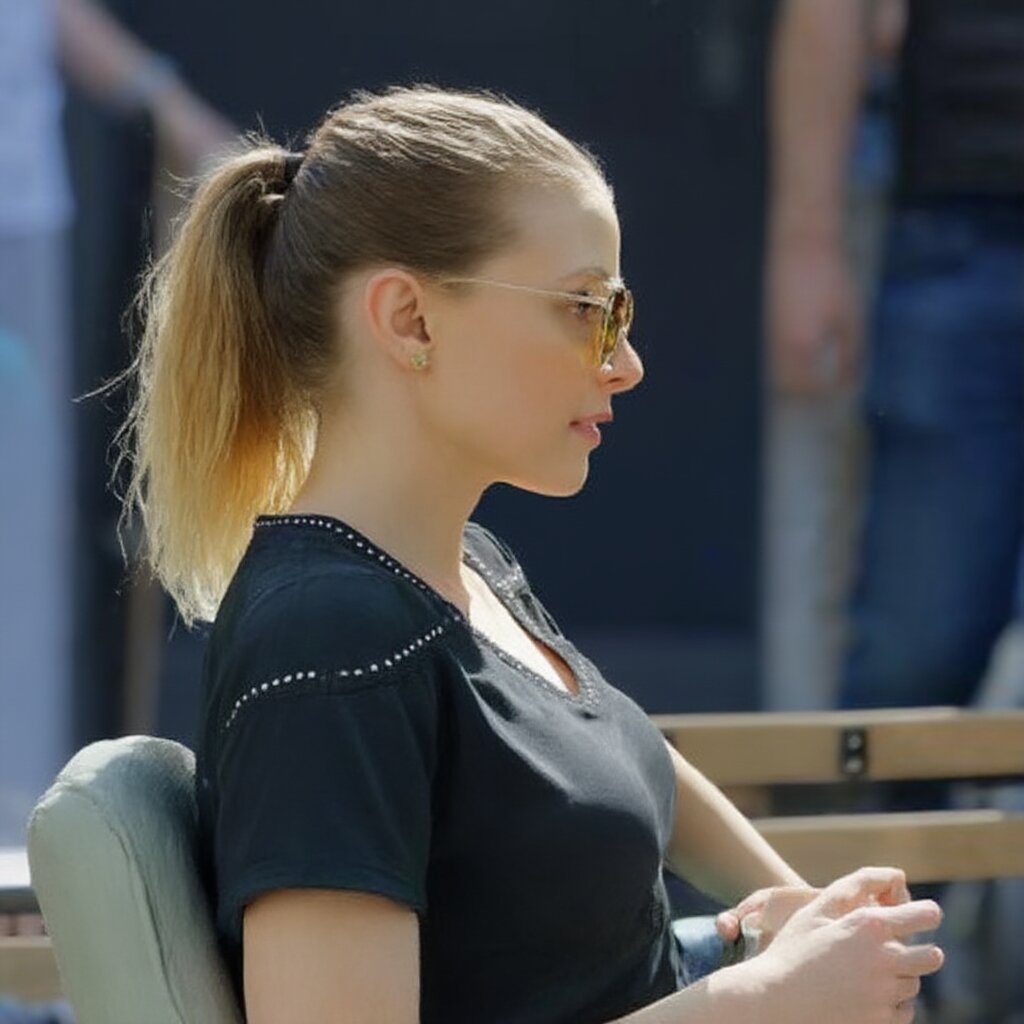} &

        \includegraphics[width=0.16\linewidth]{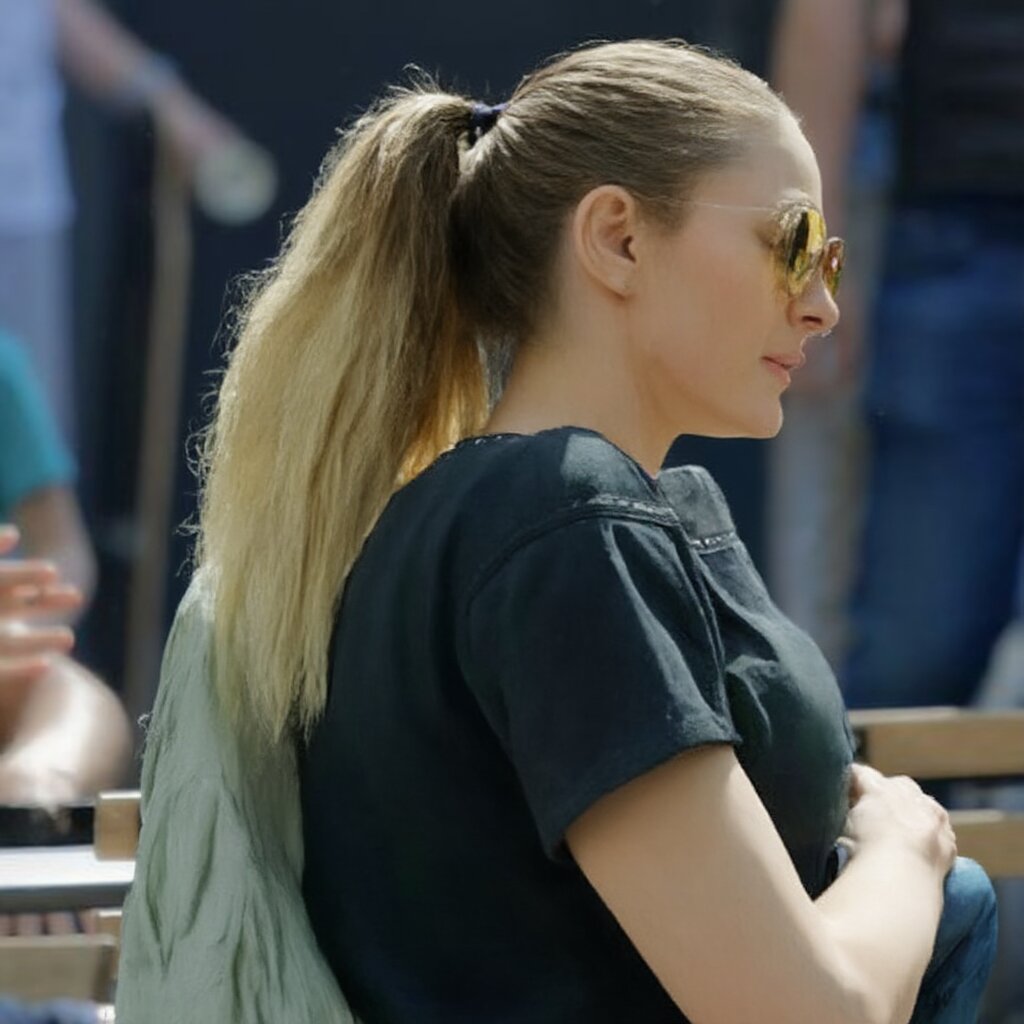} &
        
        \includegraphics[width=0.16\linewidth]{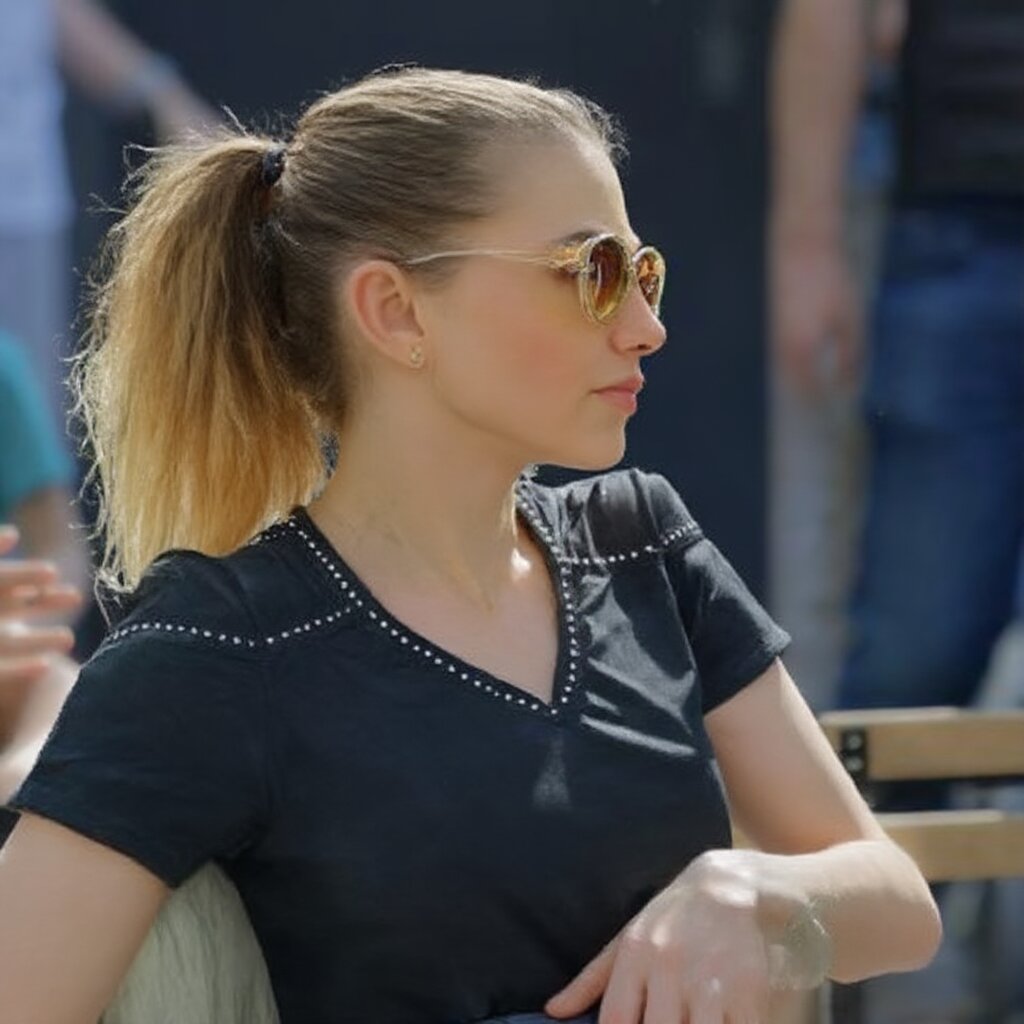} \\
        \multicolumn{6}{c}{\scriptsize \textit{``Turn right 120.0 degrees, enlarge by a factor of 1.20, move left 0.10."}.} \\
    \end{tabular}

    \caption{\textbf{Qualitative comparison of reward settings.} Left to right: source image, SFT only, w/o rotation, w/o translation \& scaling, w/o image similarity, and ours (joint reward). Source images from PIE-Bench \cite{ju2023direct}.}
    \label{fig:reward_function}
\end{figure}

\clearpage

\end{document}